\documentclass[10pt,journal,compsoc]{IEEEtran}

\ifCLASSOPTIONcompsoc
  \usepackage[nocompress]{cite}
\else
  \usepackage{cite}
\fi
\ifCLASSINFOpdf
  \usepackage[pdftex]{graphicx}
  \graphicspath{{./figures/}}
  \DeclareGraphicsExtensions{.pdf,.jpeg,.png}
\else
  \usepackage[dvips]{graphicx}
  \graphicspath{{../eps/}}
  \DeclareGraphicsExtensions{.eps}
\fi
\usepackage{amsmath}
\usepackage{amssymb}
\usepackage{amsfonts}

\usepackage{booktabs}

\ifCLASSOPTIONcompsoc
  \usepackage[caption=false,font=footnotesize,labelfont=sf,textfont=sf]{subfig}
\else
  \usepackage[caption=false,font=footnotesize]{subfig}
\fi

\newcommand\MYhyperrefoptions{bookmarks=true,bookmarksnumbered=true,
pdfpagemode={UseOutlines},plainpages=false,pdfpagelabels=true,
colorlinks=true,linkcolor={black},citecolor={black},urlcolor={black},
pdftitle={Multi-Modal Fingerprint Presentation Attack Detection: Evaluation On A New Dataset},
pdfsubject={Biometrics},
pdfauthor={Leonidas Spinoulas},
pdfkeywords={Fingerprint Biometrics, Presentation Attack Detection, Multi-Spectral Imaging, Short-Wave Infrared, Convolutional Neural Networks}}

\ifCLASSINFOpdf
    \usepackage[\MYhyperrefoptions,pdftex]{hyperref}
\else
    \usepackage[\MYhyperrefoptions,breaklinks=true,dvips]{hyperref}
    \usepackage{breakurl}
\fi

\usepackage{color,soul}
\usepackage{balance}
\usepackage{multirow}
\usepackage{multicol}

\renewcommand{\eqref}[1]{{{Eq. \ref{#1}}}}
\newcommand{\figref}[1]{{{Fig. \ref{#1}}}}
\newcommand{\secref}[1]{{{Section \ref{#1}}}}
\newcommand{\tabref}[1]{{{Table \ref{#1}}}}

\newif\ifsubmission
\newif\ifarxiv
\submissiontrue  
\arxivtrue 

\ifsubmission
    \newcommand{\hm}[1]{}
    \newcommand{\ls}[1]{}
    \newcommand{\mo}[1]{}
\else
    \newcommand{\hm}[1]{\sethlcolor{green}\hl{#1}}
    \newcommand{\ls}[1]{{\color{red}\hl{#1}}}
    \newcommand{\mo}[1]{\sethlcolor{yellow}\hl{\textit{\textbf{MO: }}#1}}
\fi

\begin{document}

\title{Multi-Modal Fingerprint Presentation Attack Detection: Evaluation On A New Dataset}

\ifarxiv
\author{Leonidas~Spinoulas{$^{\ast}$},
        Hengameh~Mirzaalian{$^{\ast}$}, \\
        Mohamed~Hussein,~\IEEEmembership{Member,~IEEE},
        and~Wael~AbdAlmageed,~\IEEEmembership{Member,~IEEE}
\IEEEcompsocitemizethanks{\IEEEcompsocthanksitem L. Spinoulas, H. Mirzaalian, M. Hussein, and W. AbdAlmageed are with the Information Sciences Institute (University of Southern California), Marina Del Rey, CA, 90292.\protect\\
M. Hussein is also with the Faculty of Engineering, Alexandria University, Alexandria, Egypt 21544.\protect\\
E-mail: lspinoulas@isi.edu, hengameh@isi.edu, mehussein@isi.edu, wamageed@isi.edu}
\thanks{The asterisk $^{\ast}$ next to author names denotes equal contribution.}}
\fi

%

\IEEEtitleabstractindextext{%
\begin{abstract}

Fingerprint presentation attack detection is becoming an increasingly challenging problem due to the continuous advancement of attack preparation techniques, which generate realistic-looking fake fingerprint presentations. In this work, rather than relying on legacy fingerprint images, which are widely used in the community, we study the usefulness of multiple recently introduced sensing modalities. Our study covers front-illumination imaging using short-wave-infrared, near-infrared, and laser illumination; and back-illumination imaging using near-infrared light. Toward studying the effectiveness of each of these unconventional sensing modalities and their fusion for liveness detection, we conducted a comprehensive analysis using a fully convolutional deep neural network framework. Our evaluation compares different combination of the new sensing modalities to legacy data from one of our collections as well as the public LivDet2015 dataset, showing the superiority of the new sensing modalities in most cases. It also covers the cases of known and unknown attacks and the cases of intra-dataset and inter-dataset evaluations. Our results indicate that the power of our approach stems from the nature of the captured data rather than the employed classification framework, which justifies the extra cost for hardware-based (or hybrid) solutions. We plan to publicly release one of our dataset collections.
\end{abstract}

}

\maketitle

\IEEEdisplaynontitleabstractindextext

%
\IEEEpeerreviewmaketitle

\IEEEraisesectionheading{\section{Introduction}\label{sec:Introduction}}
\IEEEPARstart{B}{iometric} authentication systems provide additional security and convenience as well as reduced cost, compared to conventional authentication methods. As a result, their use is widespread in different application domains, including law-enforcement or border and access control, for government, corporate or personal purposes. Nevertheless, such systems can be vulnerable to different types of attacks targeting different points of the underlying authentication pipeline. Arguably, the most vulnerable component of a biometric authentication system is the biometric sensor itself, due to the public accessibility of sensors, in many cases. An attack on a biometric sensor typically constitutes the presentation of a fake sample in order to either (1) impersonate a legitimate user or (2) conceal the true identity of a black-listed one. Automatic detection of this type of \emph{presentation attack} (PA) has attracted significant research interest with a myriad of \emph{presentation attack detection} (PAD) methods applied on different biometric modalities, such as fingerprint, iris or face~\cite{Marcel2019:Handbook}. However, due to the continuous advent of realistic \textit{presentation attack instruments} (PAIs), PAD is still an increasingly challenging problem. 

Fingerprint is perhaps the first modality to be used for biometric authentication, and hence, has been thoroughly studied by the biometrics and computer vision communities~\cite{JAIN2016:50Years}. Despite its wide acceptability as a universal, distinctive, and permanent biometric characteristic~\cite{Ross07:BioRecOverview}, presentation attacks have been shown to successfully spoof fingerprint authentication systems~\cite{Galbally2007:Vulnerabilities, Bowden-Peters2012:FoolingCapacitive}. As a result, significant research work has been devoted to address the problem of \emph{fingerprint presentation attack detection} (FPAD)~\cite{Marasco2014:SurveyPADFinger,Sousedik2014:PADFingerSurvey}.

FPAD methods can be categorized into \emph{software-only} or \emph{hybrid}, based on the components they add to the biometric authentication system. \emph{Software-only} or \emph{software-based} techniques, which are the most abundant in the literature, only add a software module to augment existing fingerprint authentication systems with PAD functionality. Hence, they solely depend on the data used for enrollment and recognition to perform FPAD. Examples of \emph{software-based} techniques include~\cite{Nikam2008:Curvelet, Zaghetto2017:TextureANN, TAN2010:RidgeSignalValleyNoise, Memon2011:PoreDet, GALBALLY2012:LivDetQualityBased, Antonelli2006:SkinDistortion, Abhyankar2009:WaveletPerspiration}. On the contrary, \emph{hybrid} or \emph{hardware-based} techniques employ additional hardware for FPAD along with the hardware used for fingerprint sensing. We refer to them as \emph{hybrid} techniques since they still involve software modules that process the data captured by the additional sensing hardware to deliver FPAD functionality. Examples of \emph{hybrid} techniques include~\cite{Reddy2007:PulseOximetry, Baldisserra2006:Odor, Cheng2006:OCT}.

\begin{figure}[!t]
\centering
\includegraphics[]{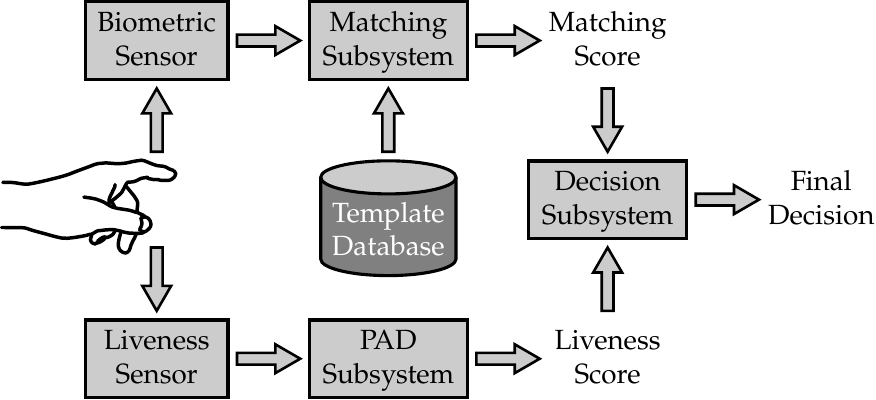}
\caption{A biometric authentication system with a hybrid fingerprint presentation attack detection (FPAD) subsystem.}
\label{fig:bio_auth_hw_pad}
\end{figure}

\emph{Software-only} FPAD techniques have attracted more interest in the research community owing to their cost-effectiveness and direct applicability on publicly available datasets. Nonetheless, we argue that \emph{hybrid} techniques should gain more attention for the following reasons:
\begin{itemize}
    \item Authentication and PAD are two fundamentally different problems. Hence, restricting them to rely on the same sensing hardware limits the progress that could be attained in each.
    \item With the continuous evolution of sophisticated PA techniques and the attackers' deeper understanding of the intrinsics of biometric authentication, it is becoming increasingly challenging to rely on a single sensing technology for both authentication and PAD. In fact, it has been shown that each of the major fingerprint sensing technologies (e.g., optical, capacitive or thermal) is vulnerable to at least one type of material used for PAIs (e.g., gelatin, silicone or glycerin)~\cite{vanderPutte2000:DontBurn, Matsumoto2002:Gummy, Barral2009:Glycerin, Barral2010:Thesis:Biometrics}.
    \item Not only attack fabrication technology improves but also the effect of attacks is becoming more devastating, especially when a single successful attack can be used to impersonate multiple individuals at the same time~\cite{Roy2018:EvolMasterPrint, Bontrager2018:DeepMasterPrints}. Consequently, especially for security critical applications, the additional hardware cost of \emph{hybrid} approaches is well justified.
    \item \emph{Hybrid} PAD techniques can enhance the reliability of a biometric authentication system. Consider the system illustrated in \figref{fig:bio_auth_hw_pad} which employs a hybrid PAD subsystem with a parallel matching branch (similar to the one introduced in~\cite{Johnson2014:EvaluationOfPAD}). In this design, the separation between the matching and PAD pipelines
    enhances security since an attacker would have to simultaneously compromise both processing pipelines to succeed. Such task is more challenging compared to the single-point attacks typically found in \emph{software-only} PAD approaches~\cite{Ratha01:EnhancingSecurity}.
\end{itemize}

\begin{figure}[!t]
\centering
\subfloat[Front-illumination.]{\includegraphics[width=0.399\columnwidth]{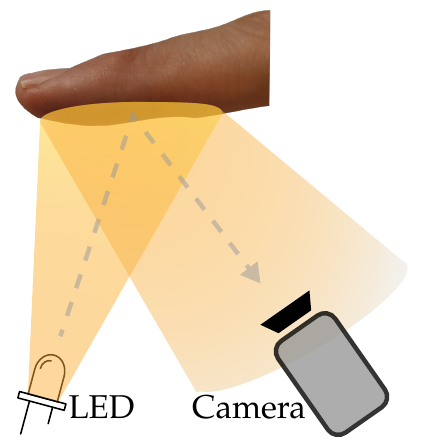}}
\hspace{0.35cm}
\subfloat[Back-illumination.]{\includegraphics[width=0.399\columnwidth]{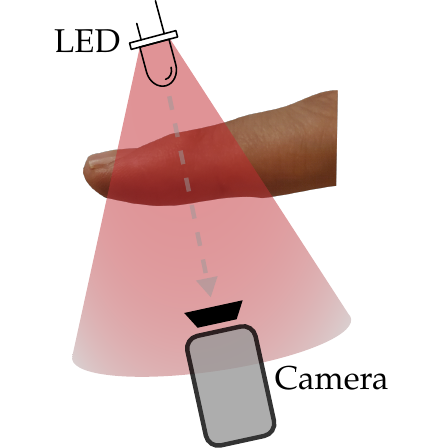}}
\caption{Front vs. Back Illumination}
\label{fig:illumination_types}
\end{figure}

In this paper, we focus on \emph{hybrid} techniques and evaluate a number of recently introduced sensing modalities for FPAD against legacy data from fingerprint authentication systems. Specifically, we investigate the performance of a novel fully-convolutional neural network (FCN) model for FPAD on images captured under different illumination conditions, namely: 
\begin{enumerate}
    \item \emph{Visible} (VIS) and \emph{near-infrared} (NIR), denoted together as $F_{M}$. 
    \item \emph{Short-wave-infrared} (SWIR), denoted as $F_S$.
    \item \emph{Laser speckle contrast imaging} (LSCI), denoted as $F_L$.
    \item \emph{Near-infrared back-illumination}, denoted as $B_N$.
\end{enumerate}
In this notation, the letter $F$ stands for front-illumination-based sensing, in which the illumination source and the camera are on the same side with respect to the finger; and the letter $B$ stands for back-illumination-based sensing, in which the illumination source and the camera are on two opposite sides of the finger. The two types of illumination are illustrated in \figref{fig:illumination_types}. The subscripts in the aforementioned notation refer to the type of illumination used. To assess the value of \emph{hybrid} techniques, the performance of our FPAD model using these modalities -- individually or in different combinations with one another -- is compared to the performance of the same model on legacy fingerprint images, used in \emph{software-only} techniques. We will refer to the unconventional sensing modalities data as \textit{prototype data}.

The performance evaluation in this paper is conducted over a new large dataset, named Presentation Attack Detection from Information Sciences Institute (\emph{PADISI}). \emph{PADISI} includes data from three different biometric modalities: fingerprint, face, and iris that were collected using the developed system in~\cite{Spinoulas2020}. In this paper, we only study the fingerprint portion, which we will refer to as \emph{PADISI-Finger}. \emph{PADISI} was collected at two different sites. The first site is the University Park Campus of the University of Southern California (\emph{USC}), in Los Angeles, California while the second is in Columbia, Maryland in a facility of the Applied Physics Laboratory (\emph{APL}) of John Hopkins University. The data at the two sites were collected using two different replicas of the system in~\cite{Spinoulas2020}.  Additionally, the \emph{APL} collection included legacy fingerprint data collected via a number of commercial fingerprint sensors. Most of the analysis in this paper is done on the \emph{USC} data. This dataset will become publicly available upon the acceptance of this manuscript. The \emph{APL} collection is used for inter-site evaluation and for comparison with legacy data. The public release of the \emph{APL} data is beyond the control of the authors in this work.

The work presented in this paper builds on top of our prior work~\cite{wifs:Hussein:2018, LSCI_FPAD2019}, with the following additional notable contributions:
\begin{itemize}
    \item This is the first study that analyzes the PAD capabilities of unconventional sensing modalities for fingerprint compared to legacy data. Our results highlight the power of the studied sensing modalities in tackling different types of attacks.
    \item The SWIR and LSCI data in the new datasets have significantly higher resolution ($\sim 192$ ppi -- pixels per inch) compared to the data used in our prior work~\cite{wifs:Hussein:2018, LSCI_FPAD2019}. The resolution in the old data was only $\sim 35$ for SWIR and $\sim 135$ for LSCI.  
    \item Our evaluation covers each unconventional modality individually as well as their combinations and compares all that to the performance on legacy data. Such thorough analysis provides valuable information to practitioners on the complementary nature among groups of sensing modalities and their power in comparison to legacy data.
    \item We employ a novel FCN model, which is more efficient and more powerful than our prior patch-based convolutional neural network (CNN) models~\cite{wifs:Hussein:2018, LSCI_FPAD2019}. To validate the effectiveness of this model, we present its evaluation on the LivDet$2015$ dataset~\cite{LivDet2015} achieving superior performance to state-of-the-art.
    \item Our evaluation protocols cover a wide spectrum of scenarios. We use $3$-fold cross-validation on the USC collection to assess the power of different sensing modalities. Additionally, we follow an inter-collection evaluation (training on \emph{USC} and testing on \emph{APL} data) to assess the effect of changes in demographic and PAI distributions. Finally, we follow a leave-one-attack-category out cross-validation to assess the performance on completely unseen attacks.
    \item The data collected at \emph{USC} will be publicly released upon the acceptance of this manuscript, which constitutes a valuable asset to the research community. To our knowledge, this will be the first public dataset covering such a broad range of sensing modalities for FPAD. The dataset is relatively large and covers a wide range of PAI species.
\end{itemize}
\subsection{Related Work}
\label{sec:rel_work}

Aside from \textit{alteration detection}, in which the focus is on detecting a physical alteration to a real finger for the purpose of hiding real identity, FPAD methods typically attempt to find characteristics in the input presentation that can distinguish a live finger from a dead or fake one~\cite{Sousedik2014:PADFingerSurvey}. This explains why these methods are collectively referred to as \textit{liveness detection}~\cite{Sousedik2014:PADFingerSurvey}. 

Numerous proposed approaches are based on physiological characteristics of the finger that are directly sensed as part of the biometric authentication system. These characteristics can be either static or dynamic. \textit{Static} characteristics, such as odor~\cite{Baldisserra2006:Odor}, skin resistance~\cite{Kulkarni:2015:surveyFPDA,Drahansky2008}, perspiration~\cite{Marasco2012PR}, and internal finger structure measured by optical coherence tomography (OCT)~\cite{Hogan2019Patent,Cheng2006:OCT,Chugh2019OCT,Shiratsuki2005}, are typically extracted from a single image of the fingerprint. Other methods exploit multiple images to extract static characteristics, such as techniques based on multi-spectral imaging (MSI)~\cite{Tolosana2018,Rowe2006Patent,Rowe2008:MSI,Maltoni:2009,HENGFOSS2011:DynamicSWIR,Drahansky2013,Tolosana2019} or multi-view imaging~\cite{Raspireader2018}. On the other hand, \textit{dynamic} characteristics are derived by nature when processing multiple fingerprint images, e.g., a time series of images to measure finger distortion and elasticity~\cite{Antonelli2006:SkinDistortion,Jia:ICB2007}, heartbeat~\cite{Abhyankar2009:WaveletPerspiration} or blood flow~\cite{Lapsley:1998,vaz:16}.

Most of the aforementioned approaches are \textit{hybrid} techniques since they involve additional hardware components for detecting certain physiological characteristics. However, the distinction between bona-fides and attacks may not only rely on the lack of liveness-related physiological attributes but also on the existence of artifacts on a fake fingerprint, stemming from its fabrication process. Consequently, most \emph{software-only} techniques attempt to differentiate between bona-fide and attack samples based on pattern recognition and signal processing techniques without making the distinction between liveness detection or artifact detection. 

Traditionally, \emph{software-only} FPAD methods apply conventional classification techniques (e.g., support vector machines - SVMs) using hand-crafted features such as wavelet and  gray level co-occurrence matrix of optical images~\cite{Nikam2008,Derakhshani:2003:PR,Tan:2006:CVPR,Nikam2008:Curvelet}, 
histogram and binarized statistical image features~\cite{Keilbach2018,Ghiani2013BTAS,Gonzalez2019ICB}, scale-invariant feature descriptors~\cite{Marta2019ICB}, Weber local descriptor~\cite{Gragnaniello2013IEEEBMSSMA},  frequency domain information~\cite{Gragnaniello2015LCP}, and pore location distribution~\cite{Marcialis2010ICPR}. 

More recently, many \emph{software-only} FPAD approaches have been proposed utilizing CNNs~\cite{Pala2017ACVPR,Nogueira:2016:IEEETIFS,Gajawada2019ICB,Menotti2015}. Nogueira et al.~\cite{Nogueira:2016:IEEETIFS}  fine-tuned  AlexNet~\cite{AlexNet:NIPS2012} and VGG~\cite{VGG:Simonyan14c} architectures to perform liveness detection of fingerprints. A classical CNN consisting of four $2$D convolutional layers with a binary cross-entropy loss was used by Wang et al.~\cite{Wang:2015:ChineseConf}. Bhanu et al.~\cite{Bhanu:2017:DL_bio_Book} used triplet loss in their network to minimize the intra-class distances of the patches belonging to the same class while maximizing the inter-class distances. Chugh et al.~\cite{Chugh2017,AnilICB2019,Chugh2017IJCB} used MobileNet-v1 over the centered and aligned patches extracted around fingerprint minutiae to discriminate between fake and real fingerprints of optical images. Park et al.~\cite{Eunsoo:2018} included fire and gram modules within their network to learn the textures of bona-fide and PA samples. Kim et al.~\cite{KIM:2016} employed deep belief networks and used contrastive divergence for FPAD.

There exist few CNN-based \emph{hybrid} techniques for FPAD. For instance, recently, CNN models were used with LSCI~\cite{wifs:Hussein:2018,LSCI_FPAD2019,Kolberg2019BIOSIG} and 
SWIR~\cite{Tolosana2018, Tolosana2019} imaging. MSI-based \emph{hybrid} techniques are particularly relevant to our research. MSI can reveal distinguishing characteristics of real human skin compared to a multitude of materials used to create fingerprint PAs. In the simplest form of MSI, multiple images of a given target are captured while the target is actively illuminated by different wavelengths in each frame. Each wavelength exhibits varying penetration, absorption and reflection properties for different materials. These phenomena can be utilized in distinguishing real human skin from other materials. Row et al. first introduced MSI to fingerprint image acquisition in~\cite{Rowe2008:MSI} but all wavelengths were in the visible spectrum; specifically $430$nm-$630$nm and white. Response to visible spectrum illumination significantly varies between different skin tones, which limits its utility for FPAD. Very limited applications of MSI in FPAD beyond the visible domain exist, such as the work on the visible and near infrared spectrum (i.e., $400$nm-$850$nm)~\cite{Mao2011:MSI-NIR}, and on the visible to SWIR regimes ($400$nm-$1650$nm), whose dynamic characteristics were investigated as a means of distinguishing live fingers from cadavers~\cite{HENGFOSS2011:DynamicSWIR}. In both~\cite{Mao2011:MSI-NIR} and~\cite{HENGFOSS2011:DynamicSWIR}, very few samples were used. In this paper, we present the most comprehensive study of MSI on FPAD. Not only, our study covers wavelengths in the broad range VIS-SWIR, but also includes front, back, LED-based and laser-based illumination. Furthermore, we use relatively large datasets containing a wide variety of attacks.

\begin{table*}[!t]
    \caption{Overview of collected data per finger by the utilized finger biometric sensor suite.}
    \label{tab:finger_data_summary}
    \centering
    \resizebox{\linewidth}{!}{
    \begin{tabular}{c||ccccccccccccc}
         \toprule \hline
         \textbf{Camera} & \multicolumn{8}{c||}{VIS/NIR~\cite{basler_nir_finger} - $1282 \times 1026$ pixels - $12$ bits (stored as $16$)} & \multicolumn{5}{c}{SWIR~\cite{bobcat} - $320 \times 256$ pixels - $16$ bits} \\ \hline
         
         \textbf{Sensing Modality} & \multicolumn{7}{c}{$F_{M}$} & \multicolumn{1}{||c||}{$B_{N}$} & 
         \multicolumn{4}{c||}{$F_{S}$} &
         $F_{L}$ \\ \hline
         
         \textbf{Illumination} & white & $465$nm & $591$nm & $720$nm & $780$nm & $870$nm & 
         $940$nm & \multicolumn{1}{||c||}{$940$nm} & $1200$nm & $1300$nm & $1450$nm & 
         \multicolumn{1}{c||}{$1550$nm} & $1310$nm~\cite{laser} \\ \hline
         
         \textbf{Illuminated} & 
         \multirow{2}{*}{$1$} & 
         \multirow{2}{*}{$1$} & 
         \multirow{2}{*}{$1$} & 
         \multirow{2}{*}{$1$} & 
         \multirow{2}{*}{$1$} & 
         \multirow{2}{*}{$1$} & 
         \multirow{2}{*}{$1$} & 
         \multicolumn{1}{||c||}{\multirow{2}{*}{$20$}} & 
         \multirow{2}{*}{$1$} & 
         \multirow{2}{*}{$1$} & 
         \multirow{2}{*}{$1$} & 
         \multicolumn{1}{c||}{\multirow{2}{*}{$1$}} & 
         \multirow{2}{*}{$100$} \\
         \textbf{Frames} & & & & & & & & \multicolumn{1}{||c||}{} 
         & & & & \multicolumn{1}{c||}{} & \\ \hline
         
         \textbf{Non-Illuminated} & 
         \multirow{2}{*}{$1$} & 
         \multirow{2}{*}{$1$} & 
         \multirow{2}{*}{$1$} & 
         \multirow{2}{*}{$1$} & 
         \multirow{2}{*}{$1$} & 
         \multirow{2}{*}{$1$} & 
         \multirow{2}{*}{$1$} & 
         \multicolumn{1}{||c||}{\multirow{2}{*}{$-$}} &
         \multirow{2}{*}{$4$} & 
         \multirow{2}{*}{$4$} & 
         \multirow{2}{*}{$4$} & 
         \multicolumn{1}{c||}{\multirow{2}{*}{$4$}} & 
         \multirow{2}{*}{$-$} \\
         \textbf{Frames} & & & & & & & & \multicolumn{1}{||c||}{} 
         & & & & \multicolumn{1}{c||}{} & \\ \hline \bottomrule
    \end{tabular}}
\end{table*}

\section{Finger Biometrics System Design}
\label{sec:finger_station}
The utilized finger biometrics sensor suite has been improved and simplified compared to its previous version in~\cite{wifs:Hussein:2018}. In this section, we provide a brief overview of the system focusing on the rationale for using the selected sensing modalities, supported by relevant literature. The interested reader can refer to~\cite{Spinoulas2020} for more technical details about the design, selected hardware components and its operation. The schematic diagram of the system is shown in \figref{fig:finger_biometric_suite}. A finger is placed on the slit and is illuminated from the bottom side (front-illumination) by multi-spectral LEDs and a laser as well as the top side (back-illumination) by NIR LEDs while being observed by two cameras from a distance of $\sim 35$cm. Each camera is sensitive to the VIS/NIR or the SWIR spectrum, respectively. The sensor is touch-less in the sense that no platen covers the finger slit.

Each chosen sensing modality in our station provides indicative information for either the liveness of a presentation or the presence of a fabrication artifact, as follows:
\begin{itemize}
    \item \underline{Front VIS/NIR illumination ($F_{M}$) data}: NIR multi-spectral images can provide ample information about the spectral and textural characteristics of the material of a presentation~\cite{Mao2011:MSI-NIR}. Further, they can be used as legacy compatible data, as shown in~\cite{Spinoulas2020}.

    \item \underline{Front SWIR illumination ($F_{S}$) data}:
    Human skin has a distinctive response in the SWIR spectrum that is independent of skin tone. In a recent study, conducted by the National Institute of Standards and Technology (NIST), the variability of skin response due to differences among people was found to be less significant than the variability due to instrument characteristics only beyond the $1100$nm wavelength~\cite{Cooksey2014:SkinReflectance}. The same observation was confirmed in a successful application of PAD using MSI in the SWIR domain, for face biometrics~\cite{Steiner2016:MSI-SWIRFace}, and more recently, using the previous version of our data, on fingerprint biometrics~\cite{GomezBarrero2018TowardsFP,wifs:Hussein:2018,Tolosana2018,Tolosana2019,GomezBarrero:NISK2019}.
    
\begin{figure}[!t]
\centering
\ifarxiv
\includegraphics[]{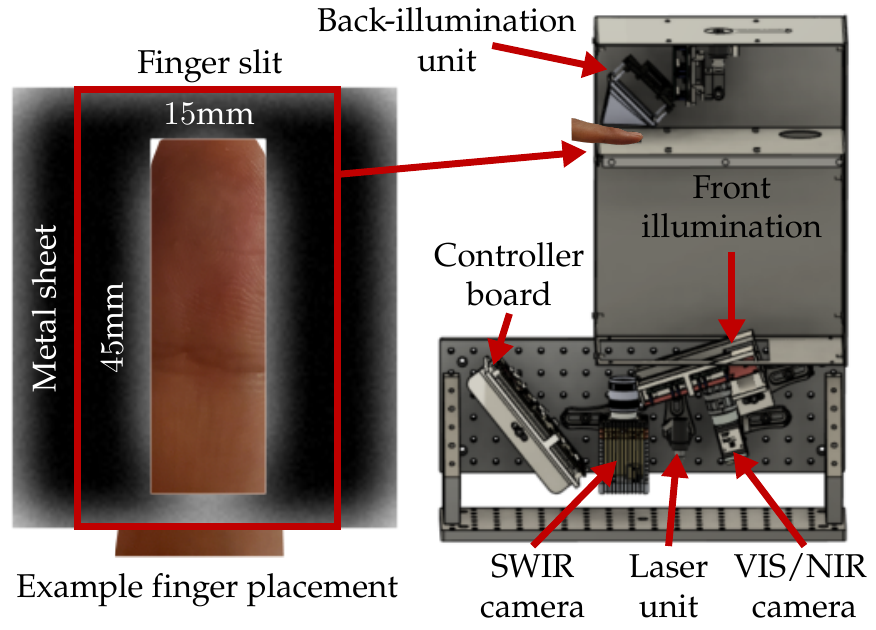}
\fi
\caption{Finger biometric sensor suite. See~\cite{Spinoulas2020} for more technical details.}
\label{fig:finger_biometric_suite}
\end{figure} 

    \item \underline{Front LSCI ($F_{L}$) data}: When laser light illuminates a surface, a random interference pattern, known as the speckle pattern, is formed. The pattern is affected by the roughness and/or temperature of the surface and appears static for stationary objects. However, the pattern changes over time when there is motion on the illuminated object, such as the movement of blood cells under the skin surface. In fact, blood perfusion in a tissue can be visualized~\cite{LSCI2013Briers} using LSCI by collecting a sequence of images. Therefore, LSCI measurements can constitute a useful liveness signal for FPAD. LSCI has attracted very little attention in the FPAD literature despite its interesting properties. Chatterjee et al.~\cite{Chatterjee2017:antispoof} conducted a study on the hardware and physics of LSCI and showed that there is significant difference between the biospeckle patterns of bona-fide and fake fingers. However, the study involved a very small dataset and did not actually evaluate FPAD performance. Using data from the previous version of our system~\cite{wifs:Hussein:2018}, Keilbach et al.~\cite{Keilbach2018,Kolberg2019BIOSIG} performed LSCI-based FPAD by applying classical classification algorithms (e.g., SVMs), on a set of hand crafted features, such as intensity histograms or LBP features. Later on, Hussein et al.~\cite{wifs:Hussein:2018} and Mirzaalian et al.~\cite{LSCI_FPAD2019} utilized various spatio-temporal neural networks to perform LSCI-based FPAD which demonstrated promising performance.

    \item \underline{Back NIR illumination ($B_{N}$) data}: The internal vascular pattern of a hand can be visualized using NIR illumination~\cite{Mao2011:MSI-NIR,Badawi2006HandVB}. In the presence of a finger between a light source and a camera, the collected image intensities represent the amount of light penetrating through the sample, as compared to front-illumination which measures the reflected light, instead~\cite{Badawi2006HandVB}. Vessel structures appear darker since light is absorbed by the hemoglobin in the blood and can be used as an indicator of liveness for FPAD. Back-illuminated NIR images from our previous dataset were used for FPAD in~\cite{ChristophBook2020}.
\end{itemize}

\begin{figure*}[!t]
\centering
\includegraphics[]{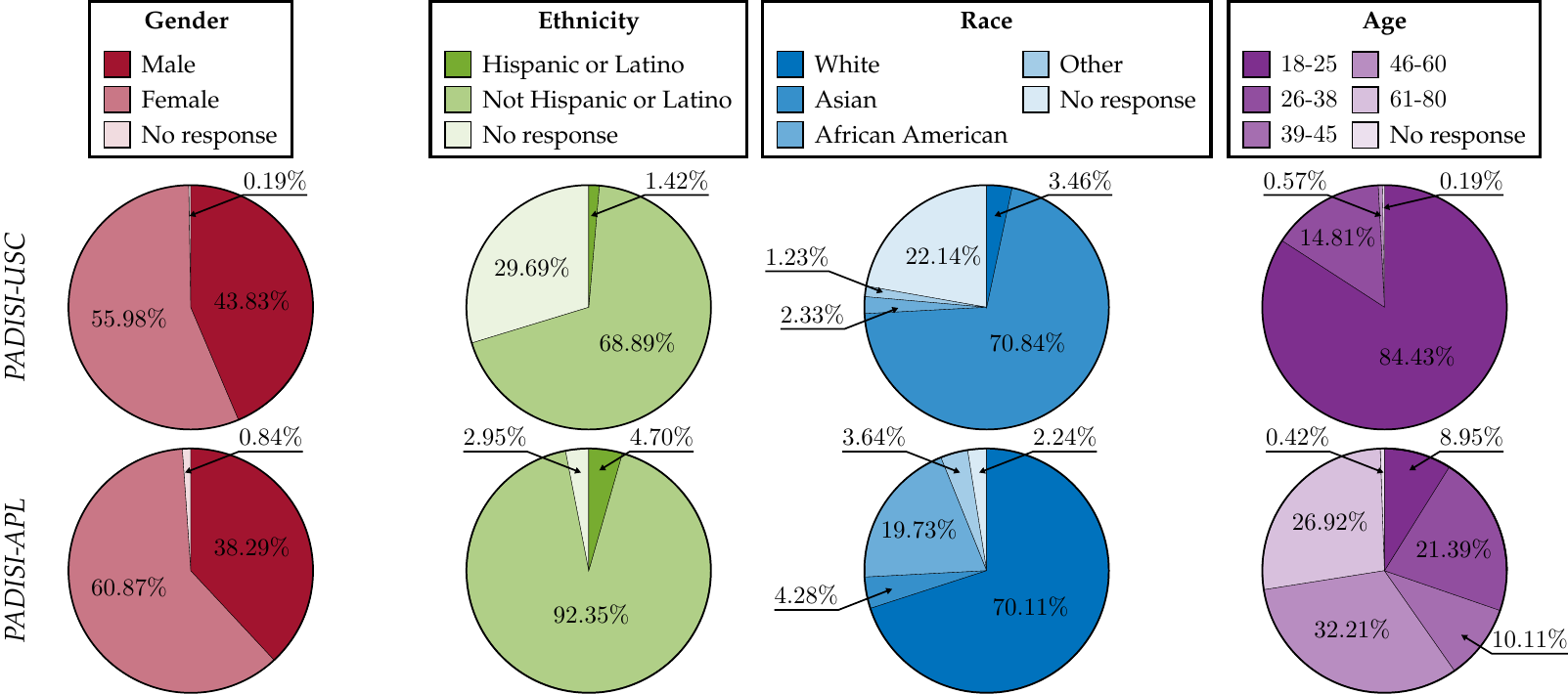}
\caption{Demographic information for the collected datasets. A summary of the collected samples per dataset can be found in~\tabref{tab:padisi_datasets}.}
\label{fig:dataset_stats}
\end{figure*}

\begin{figure*}[!t]
\centering
\ifarxiv
\includegraphics[]{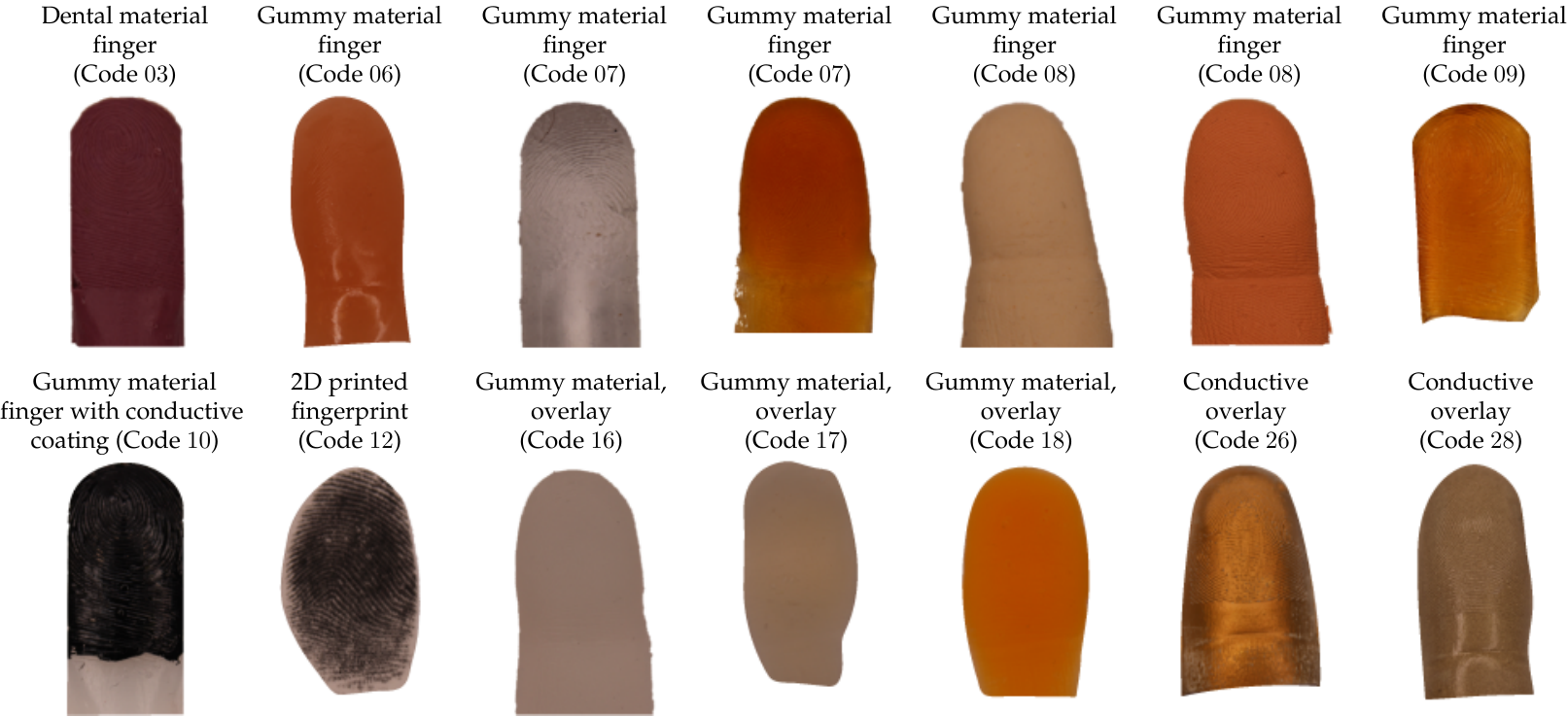}
\fi
\caption{Images of select PAIs used in the \emph{PADISI-USC} data collection. More details for each PAI can be found in~\tabref{tab:PAI_information}.}
\label{fig:finger_pais}
\end{figure*}

 It is worth noting that select combinations of the sensing modalities were already studied on the previous version of our dataset, as in~\cite{GomezBarrero:NISK2019, Marta2019ICB, MartaGomez:SITIS2018,wifs:Hussein:2018}. In this study, we evaluate FPAD performance on our new dataset using a comprehensive list of experiments.
 
 A summary of the captured data for a single finger is presented in \tabref{tab:finger_data_summary}. In multiple cases, data is also captured without active illumination and such images can be used as reference frames for ambient illumination conditions. We developed two replicas of our sensor suite each one used at each collection site (\emph{USC} and \emph{APL}). Note that, despite best efforts, some system adjustments could not be identical in the two replicas, such as lens shutter adjustments. Also, despite the closed-box design, ambient light might leak into the station and partially influence imaging conditions.

\begin{table*}[!t]
    \centering
    \caption{PAI counts in the collected datasets. For each PAI code, we provide a general PAI description, the number of different species per dataset as well as the attributes used for grouping PAI codes in terms of material, species, transparency, and attack type. PAI categories whose appearance depends heavily on the participant and preparation method are marked with $^*$. A summary of the collected samples per dataset can be found in~\tabref{tab:padisi_datasets}. Sponsor approval is required to release additional information about each PAI code.}
    \label{tab:PAI_information}
    \includegraphics[scale=0.895]{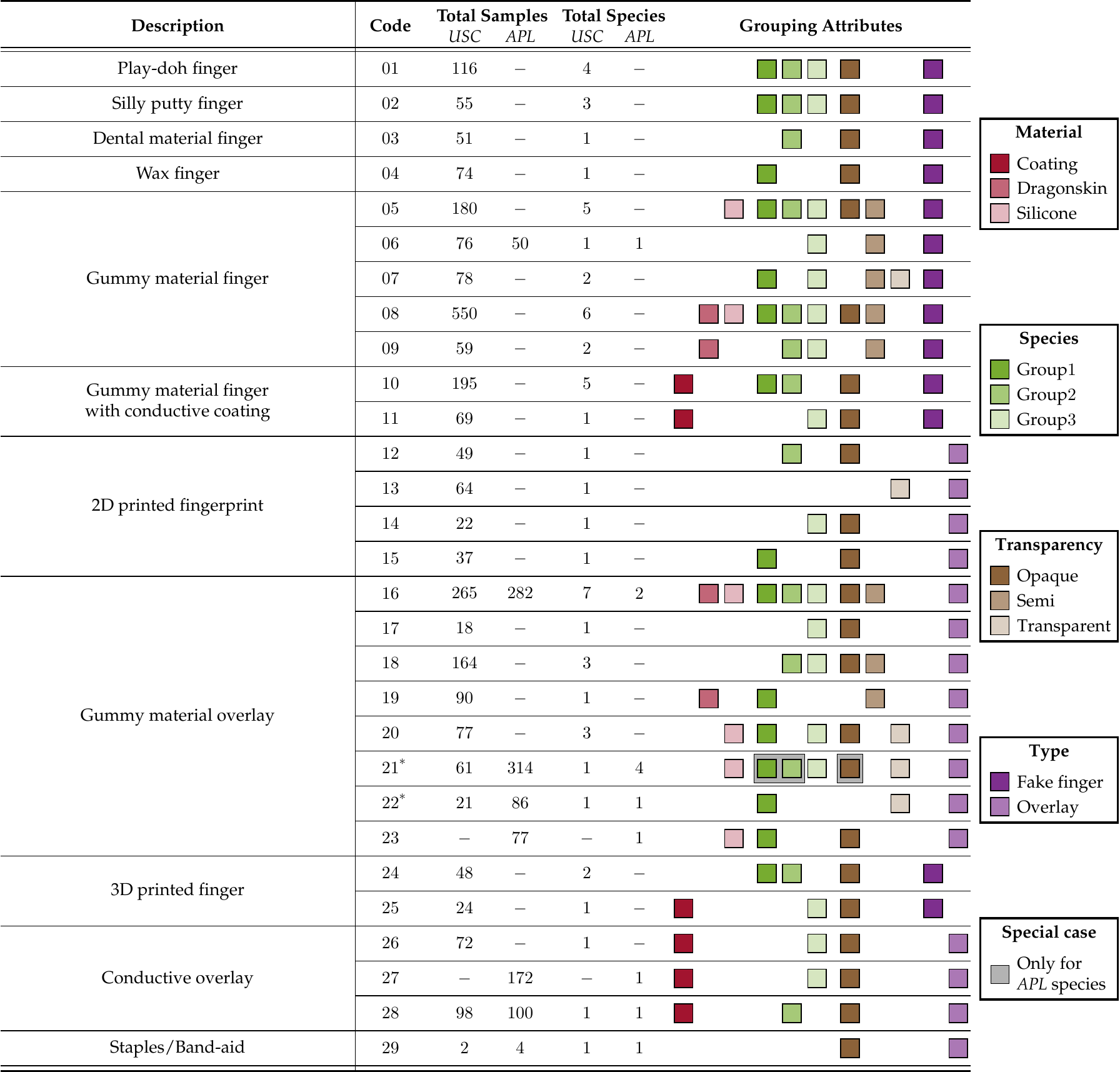}
\end{table*}

\section{Dataset}
\label{sec:dataset}
As discussed in \secref{sec:Introduction}, the \emph{PADISI} dataset was collected at two different sites (\emph{USC} and \emph{APL}). The data at \emph{USC} were collected in two separate sessions, referred to as \emph{USC-1} and \emph{USC-2}, between which minor adjustments were applied on the system. During all collections, a participant presented $4$ fingers from each hand (excluding little fingers). Each participant passed once or twice by the collection station either in the absence of any attack or in the presence of up to three attached PAIs to the
fingers of one or both hands, while a few participants participated in multiple collection sessions. At the \emph{APL} data collection, data was also collected from all participants using a series of commercial legacy sensors (see~\cite{Spinoulas2020} for details). In this work, for comparison purposes, we chose to use data from the Optical-C sensor, since it has the highest resolution and provided the most reliable data in terms of match rates (see~\cite{Spinoulas2020}). 

\begin{figure*}[!t]
\centering
\ifarxiv
\includegraphics[]{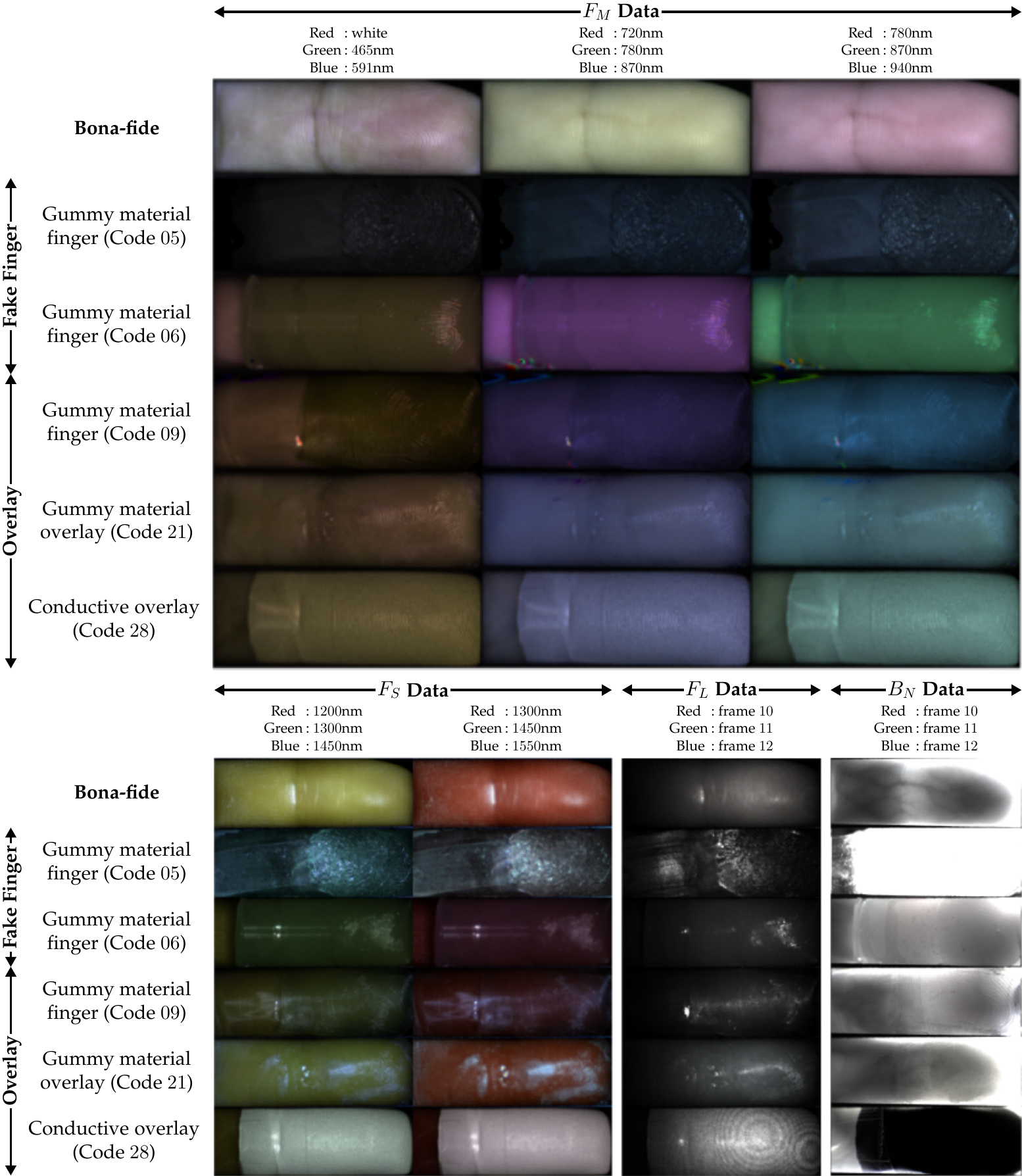}
\fi
\caption{RGB visualization of samples from the collected datasets for all types of captured data. For each image, the corresponding dark channel has been subtracted and each RGB channel has been normalized by dividing by its maximum value for visualization purposes, albeit introducing visible color artifacts in some cases. Note that, the resolution of the $B_N$ data is the same as the $F_{M}$ data but is presented smaller in the illustration.}
\label{fig:rgb_visualized_samples}
\end{figure*}

The collected data was thoroughly reviewed by the research team and samples with defects, e.g., due to finger motion or hardware failure, were excluded. A summary of the collected data counts, after these revisions, is provided in \tabref{tab:padisi_datasets} while the relevant demographic information is presented in \figref{fig:dataset_stats}. 
As observed, the \emph{APL} dataset is larger but exhibits a big imbalance between bona-fide and PAI samples. At the same time, the \emph{USC} dataset contains a much larger variety of PAI species. Moreover, there is a huge discrepancy in demographics especially in terms of age and race. 
\begin{table}[!htb]
    \centering
    \caption{Collected datasets summary. Demographic information can be found in~\figref{fig:dataset_stats} while more details about PAIs are provided in~\tabref{tab:PAI_information}.}
    \label{tab:padisi_datasets}
    \begin{tabular}{c||ccc} 
         \toprule
         & \multicolumn{3}{c}{\emph{PADISI} Datasets}  \\ 
         & \emph{USC} & \emph{APL} & \emph{APL-LEGACY}  \\ \hline \hline
         \textbf{Participants} & $355$ & $672$ & $665$ \\
         \textbf{Unique fingers} & $2490$ & $5371$ & $5308$\\
         \textbf{Total samples} & $6211$ & $11444$ & $8850$ \\
         \textbf{Bona-fide samples} & $3596$ & $10359$ & $8043$\\
         \textbf{PAI Samples} & $2615$ & $1085$ & $807$ \\
         \textbf{PAI Species} & $58$ & $12$ & $12$\\
         \hline \bottomrule
    \end{tabular}
\end{table}The \emph{USC} dataset contains mostly young people of Asian origin (since it was collected in a university environment) while the \emph{APL} dataset exhibits a skewed distribution toward white people while having a more balanced age distribution. The discrepancy in the number of participants and total samples between the \emph{APL} and \emph{APL-LEGACY} data in \tabref{tab:padisi_datasets} is due to the fact that participants did not always visit the legacy sensors multiple times due to their slow speed as well as the data revision process, which was independently performed per sensor. In this table, we present only the samples from the legacy sensor that correspond to participants and fingers available in our prototype data.

Snapshots of prepared and collected PAIs are illustrated in \figref{fig:finger_pais}. Analytic information about all PAIs used during the data collections are provided in \tabref{tab:PAI_information} where PAIs are categorized using different codes, each of which can contain one or more species (e.g., referring to a different color of a specific material or a slightly different preparation method). For each PAI code, we have provided a classification scheme using a set of grouping 
attributes based on their material, species, transparency, and attack type. Some of these attributes (e.g., transparency) were selected subjectively based on the appearance of each PAI and this classification will become important in understanding upcoming results in this work. Finally, examples of the finger area of the captured data for a bona-fide sample and select types of PAIs for all sensing modalities, described in \secref{sec:finger_station}, are provided in \figref{fig:rgb_visualized_samples}. In this illustration $3$ frames are each time stacked together to form a false-colored RGB image.  

\begin{figure*}[!t]
    \centering
    \ifarxiv
    \includegraphics[scale=0.85]{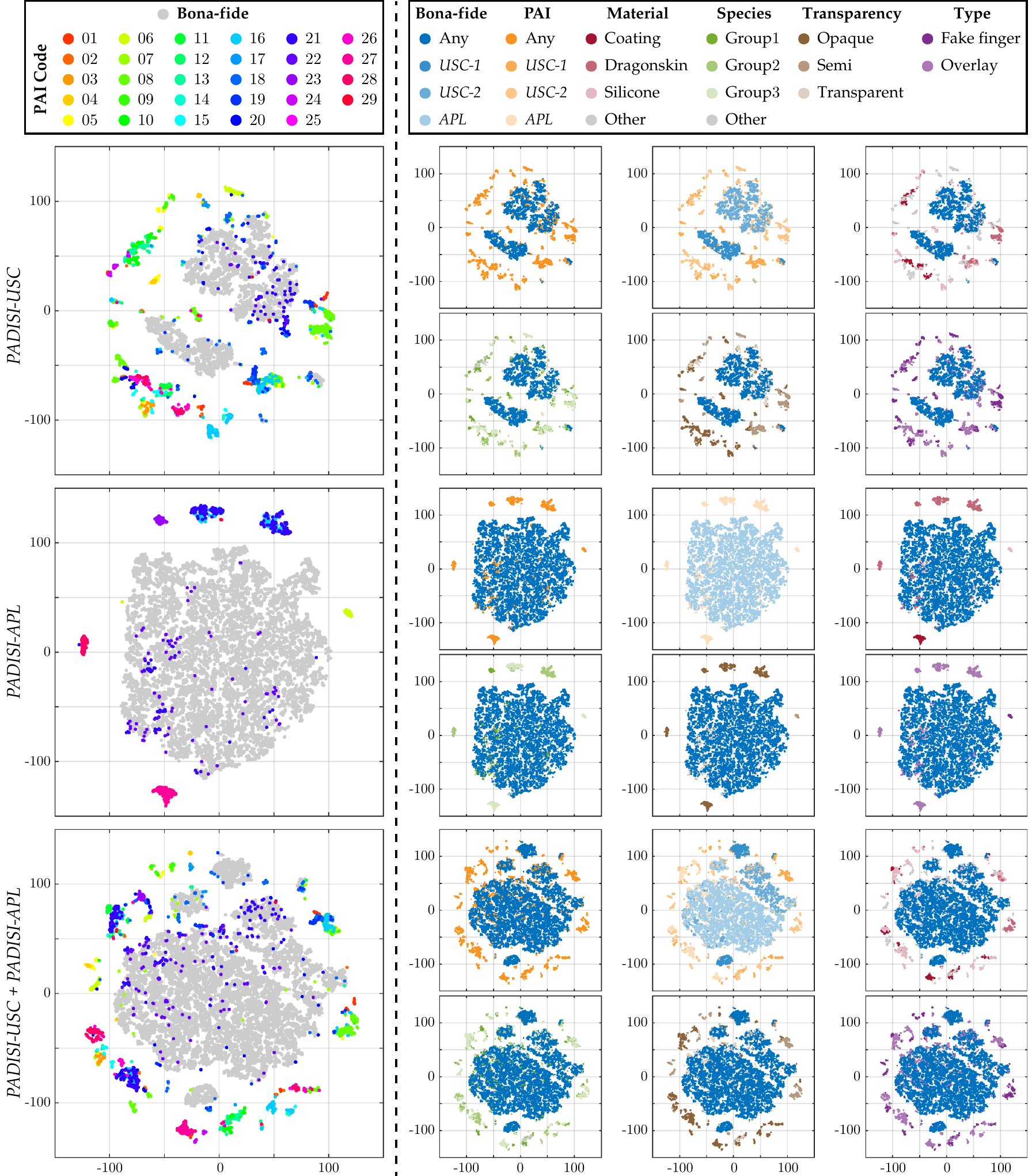}
    \fi
    \caption{t-SNE~\cite{tSNE} visualization of average intensities of $F_{M}$, $F_{S}$, $10$ frames of $F_{L}$ and $3$ frames of $B_{N}$ data for all samples in the \emph{PADISI} datasets. Visualizations are re-colored to distinguish data from different collections as well as bona-fides from PAIs with different characteristics.} 
    \label{fig:tsne_mean_intensity}
\end{figure*}

\begin{figure*}[!t]
\centering
\includegraphics[]{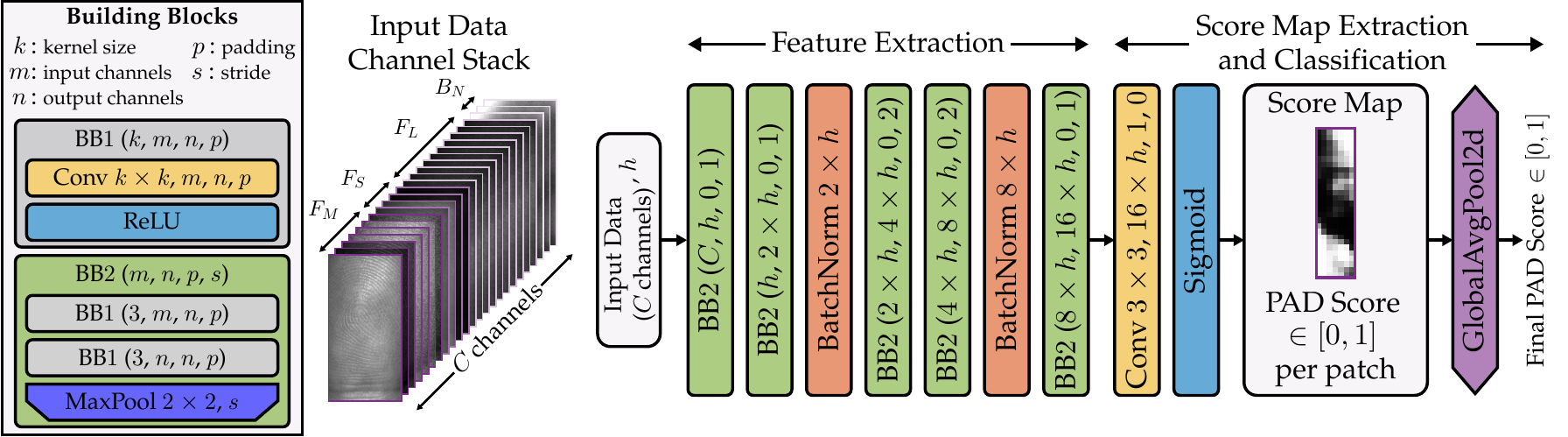}
\caption{Proposed FCN network architecture: Given parameter $h$, an input image of $C$ channels is first converted into a two dimensional score map (corresponding to a PAD score per patch) from which the final PAD score is extracted by global average pooling.} 
 \label{fig:network}
\end{figure*}

\subsection{Mean Intensity Analysis}
\label{sec:dataset_mean_intensity_analysis}
Based on the appearance of the images in \figref{fig:rgb_visualized_samples}, one might argue that the average intensities of each spectral channel of the captured data could be sufficient for obtaining a reasonable FPAD performance, since the images of bona-fides and PAIs look drastically different. In order to understand the characteristics of the captured multi-spectral data in more detail, we conduct an analysis of the average intensity of each channel using t-SNE visualizations~\cite{tSNE}, where we use all available $F_{M}$ and $F_{S}$ data, $10$ frames from the $F_{L}$ data and $3$ frames from the $B_{N}$ data (see \tabref{tab:finger_data_summary}). The visualizations are presented in \figref{fig:tsne_mean_intensity} where samples are re-colored to distinguish bona-fides from specific PAI codes, bona-fides and PAIs in general, bona-fides and PAIs from each data collection session as well as bona-fides and PAIs with the grouping attributes introduced in \tabref{tab:PAI_information}. From the presented $2$D distributions, we can make the following observations:
\begin{itemize}
    \item PAI codes tend to form individual clusters, supporting the use of multi-spectral data for observing the distinctive response of different PAI materials.
    \item Bona-fide samples of the \emph{USC-1} and \emph{USC-2} data collections appear to have a large separation mainly due to the adjustments in our system (as discussed earlier). At the same time, bona-fide samples between the \emph{USC} and \emph{APL} data collections also exhibit significant separation possibly because of their vastly different demographics (as shown in \figref{fig:dataset_stats}).
    \item It is apparent that a large number of PAIs becomes separable from bona-fides simply using the average intensity features, strengthening the power of multi-spectral data for FPAD. However, the analysis demonstrates that certain PAI types might be particularly hard to detect, the majority of which are transparent overlays made of multiple materials.
\end{itemize}
The presented visualization can also provide an estimate of how challenging a dataset is and could be used for assigning weights to PAI codes, based on their distance from the bona-fide cluster.

\section{Fully-Convolutional Neural Network Model}
\label{sec:method}
Deep neural networks (DNNs) have repeatedly delivered breakthroughs in multiple research disciplines. However, they are notoriously known for being data-hungry: substantial amounts of data have to be used to avoid over-fitting during training. Despite the relatively large size of our dataset, it is still small for effectively training a DNN model. This problem has been addressed in prior work on our data in two different ways. In~\cite{wifs:Hussein:2018,LSCI_FPAD2019}, patch-based models were employed by extracting hundreds of $8\times8$ patches from an input sample and classifying it by taking the average score over all patches. This approach proved very effective but has two main drawbacks. First, it involves repeated computations when processing consecutive patches, which typically overlap. Second, it does not scale well as the input image size increases, in which case sparse sampling of patches becomes necessary for the approach to be feasible. The alternative approach of transfer learning was used in~\cite{Marta2019ICB,GomezBarrero:NISK2019,Tolosana2019}, where a pre-trained network is fine-tuned on our data instead of trained from scratch. The main issue with transfer learning is relying on existing models, which might be too complex for the task at hand.

In this paper, we introduce a novel model for FPAD that avoids the drawbacks of both the patch-based and transfer learning-based approaches. Our new model uses an FCN structure that maps a whole region of interest (ROI) of the input finger image to a map of classification scores, which is then averaged to produce the overall classification score of the input sample. The details of the network structure are shown in \figref{fig:network}. The network consists of five convolutional blocks, with a batch-normalization layer after each group of two-blocks. Starting from a specified number of convolutional maps in the first block ($h$ in \figref{fig:network}), the number of convolutional maps is doubled in the transition between each two consecutive blocks. The final block is followed by a convolutional layer that produces a single map using a sigmoid activation, which represents the patch-wise classification score. Note that, there are no explicitly cropped patches in the FCN model. The \textit{patch} is an implicit concept referring to the \textit{receptive field} of each score map element. The score map is then fed to a global average pooling (GAP) layer to produce the final score. Each score map entry is in the range $[0,1]$, enforced by the sigmoid activation. Hence, the final model output also falls in $[0,1]$. 

The resulting PAD score can be interpreted as the probability of the presence of a PA. Hence, the ground truth score is set to $1$ for PA samples and $0$ for bona-fide samples. The loss function comprises of two components, one for the final classification score $L_{GAP}$, and one for the patch classification score $L_{patch}$, as:
\begin{equation}
\label{eq:total_loss}
    L = L_{GAP} + \lambda L_{patch} \enspace,
\end{equation}
where $\lambda > 0$.
Let the score map associated with an input image $x$ be $M: R^{C\times W\times H} \rightarrow R^{W_m \times H_m}$, where $C$ is the number of channels in $x$, $W\times H$ is the input $2$D image size, and $W_m \times H_m$ is the score map size. Considering the binary cross entropy (BCE) as the loss function, the two loss components can be expressed as:
\begin{align}
    \label{eq:loss_gap}
    L_{GAP}(x) & = & BCE\left(\frac{1}{W_mH_m}\sum_i^{W_m} \sum_j^{H_m} M_{ij}(x), \enspace t\right), \\
    \label{eq:loss_pixels}
    L_{patch}(x) & = & \frac{1}{W_mH_m}\sum_i^{W_m} \sum_j^{H_m} BCE\left(M_{ij}(x), \enspace t\right) \enspace ,
\end{align}
where $t \in \{0, 1\}$ is the ground truth label of the input $x$.

\begin{figure*}[!t]
    \centering
    \includegraphics[scale=0.9]{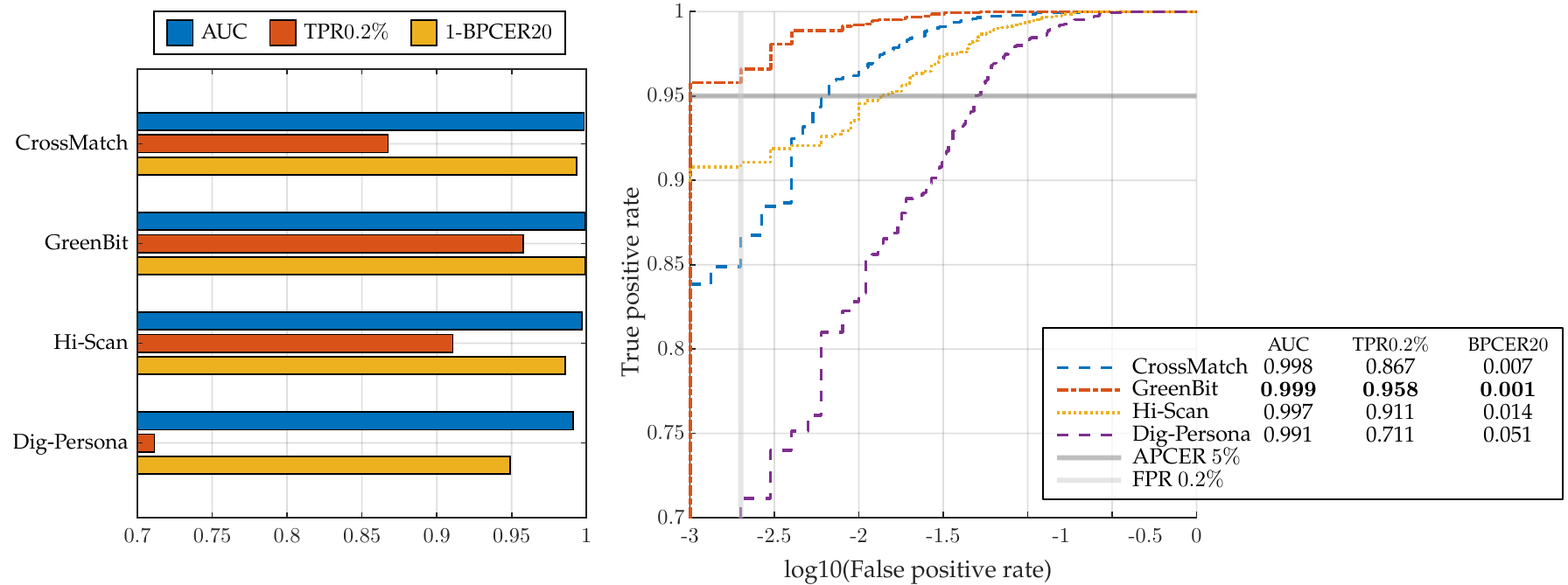}
    \caption{FPAD evaluation on the sensors of the LivDet$2015$ dataset~\cite{LivDet2015}. Left: Bar-graph visualization of the evaluation metrics; Right: ROC curves and analytic values (with best values highlighted in bold). The corresponding accuracies per sensor are presented in \tabref{tab:livdet_results}.}
    \label{fig:livdet2015_results}
\end{figure*}

For all experiments presented in this work, we use $h=16$ and $\lambda=10$, while $C$ is varying depending on the provided input image channels. Based on these parameters, the effective patch size (or receptive field) of the model is an area of $54 \times 54$ pixels for each score in $M$. Training is performed for a maximum of $100$ epochs using a batch size of $16$ and the Adam optimizer~\cite{Kingma2014}. The initial learning rate is set to $2\times 10^{-4}$ with a minimum learning rate of $1 \times 10^{-7}$ following a reduce-on-plateau by $0.5$ strategy (with patience $10$ epochs and threshold $1 \times 10^{-4}$) by monitoring the validation loss, when a validation set is available.

\section{Experimental Evaluation} 
\label{sec:results}
This section presents the experimental evaluation of this work. We first study the capabilities of the presented FCN model on an existing dataset and then run a comprehensive list of experiments on the presented dataset. The analysis uses various evaluation protocols on different combinations of input channels from the collected sensing modalities in our dataset and compares their performance with legacy data, whenever appropriate. Before proceeding with the evaluation, we summarize the data pre-processing steps and evaluation metrics used in our analysis.

\paragraph*{\textbf{Data Pre-Processing}}
\label{sec:data_preprocessing}
The data, whether prototype or legacy, are first pre-processed:
\begin{itemize}
\item \underline{Prototype Data}: From the captured data, summarized in \tabref{tab:finger_data_summary}, we use all frames from the $F_{M}$ and $F_{S}$ sensing modalities, frames $10-19$ from the $F_{L}$ data and frames $10-12$ from the $B_{N}$ data, in different combinations. If non-illuminated frames are available for any spectral channel, the time-averaged non-illuminated frame is first subtracted. The data is then normalized in $[0, 1]$ using the corresponding bit depth (see \tabref{tab:finger_data_summary}). Because of the fixed relative location of the finger slit in images captured by our equipment, it was possible to set fixed ROIs for each sensing modality. The ROI was chosen to cover most of the area of the top finger knuckle which almost always consists of skin in bona-fide samples, and PAI material in PA samples. ROIs of images of different sensing modalities are all resized to 160$\times$80 pixels using bicubic interpolation and then stacked to form the multi-spectral data cube provided to our model.
\item \underline{Legacy Data}: Legacy finger images consist of a single grayscale image where the finger ridges appear dark. We detect the finger area by first binarizing the image using thresholding, applying dilation with a circular structuring element of diameter $7$ pixels and finding the centroid of the largest connected component in the resulting binary image. The centroid is used as the center for extracting an ROI of pre-defined fixed size, for all images, depending on the average finger coverage in the images of each legacy sensor. 
\end{itemize}

\begin{table}[!t]
\centering
\caption{FPAD accuracy on the test sets of the LivDet$2015$ dataset using a threshold of $0.5$. Values of existing algorithms are taken from~\cite{LivDet2015} while the best performing algorithm is highlighted in bold. 
}  
\label{tab:livdet_results}
\resizebox{\columnwidth}{!}{
\begin{tabular}{r|ccccc} \toprule
Method & GreenBit & Hi-Scan & Persona & CrossMatch & Overall     \\ \hline \hline   
nogueira  & $95.40$ & $94.36$ & $93.72$ & $\mathbf{98.10}$  &  $95.51$ \\
unina & $95.80$ & $95.20$ & $85.44$ & $96.00$  & $93.23$ \\  
jinglian  & $94.44$ & $94.08$ & $88.16$ & $94.34$ & $92.82$ \\  
anonym & $92.24$ & $92.92$ & $87.56$ & $96.57$ & $92.51$ \\  
titanz & $91.76$ & $92.36$ & $89.04$ & $91.62$ & $91.21$ \\  
hbirkholz & $91.36$ & $93.40$ & $88.00$ & $89.93$ & $90.64$ \\  
hectorn  & $90.00$ & $88.20$ & $84.20$ & $86.94$ & $87.32$ \\  
CSI$\_$MM & $86.56$ & $87.84$ & $75.56$ & $89.99$ & $85.20$ \\  
CSI & $82.12$ & $83.20$ & $76.20$ & $88.33$ & $82.71$ \\  
COPILHA  & $72.76$  & $75.64$ & $79.96$ & $69.00$ & $74.11$ \\  
UFPE II  & $87.68$ & $71.24$ & $75.44$ & $61.16$ & $73.33$ \\  
UFPE I   & $82.56$ & $64.32$ & $78.36$ & $59.97$ & $70.82$ \\  \hline
\textbf{Proposed} & $\mathbf{98.56}$ & $\mathbf{96.80}$ & $\mathbf{94.80}$ & $\mathbf{98.10}$   & $\mathbf{97.11}$ \\  \hline \bottomrule
\end{tabular}}
\end{table}

\paragraph*{\textbf{Evaluation Metrics}}
\label{sec:Metrics}
As discussed in \secref{sec:method}, the output score of the FCN model in \figref{fig:network} represents the PA probability (in $[0,1]$) of an input sample. Therefore, using a threshold of $0.5$ is a natural choice for obtaining the FPAD binary classification and, hence computing the accuracy. However, any choice of threshold without a specific target operating point is indeed arbitrary. Hence, we opt to use metrics that do not depend on a pre-set threshold. In particular, three metrics are used to compare different models in our experiments:  i) area under the receiver operating characteristic (ROC) curve, denoted as AUC;  ii) true positive rate at $0.2\%$ false positive rate, denoted as TPR$0.2\%$, which is the primary evaluation metric used in IARPA's Odin program~\cite{odin}, through which this research has been sponsored; and iii) bona-fide presentation classification error rate (BPCER) at attack presentation classification error (APCER) of $5\%$, denoted as BPCER$20$ in the ISO standard~\cite{iso}.

\begin{figure*}[!t]
    \centering
    \includegraphics[scale=0.9]{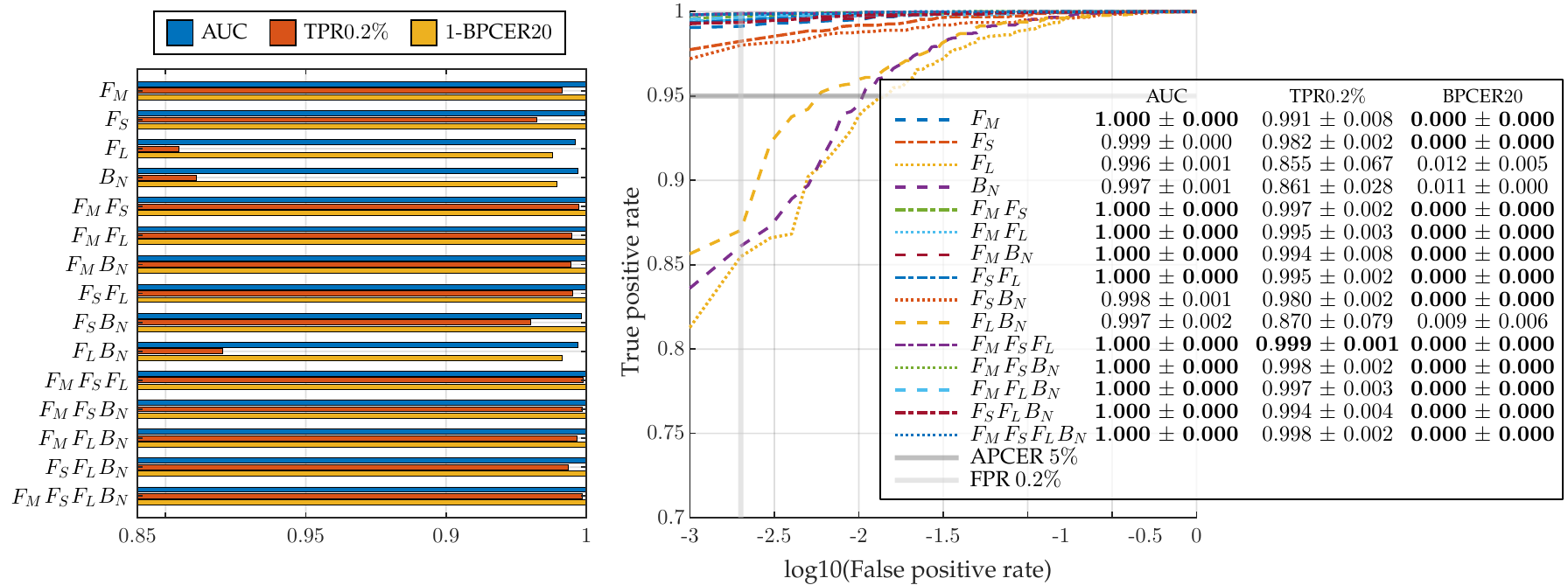}
    \caption{3FOLD FPAD evaluation on the \emph{PADISI-USC} dataset. Left: Bar-graph visualization of the evaluation metrics (mean for all folds); Right: ROC curves and analytic values (mean and standard deviation for all folds with best values highlighted in bold).}
    \label{fig:padisi_usc_3fold}
\end{figure*}

\begin{figure*}[!t]
    \centering
    \includegraphics[scale=0.9]{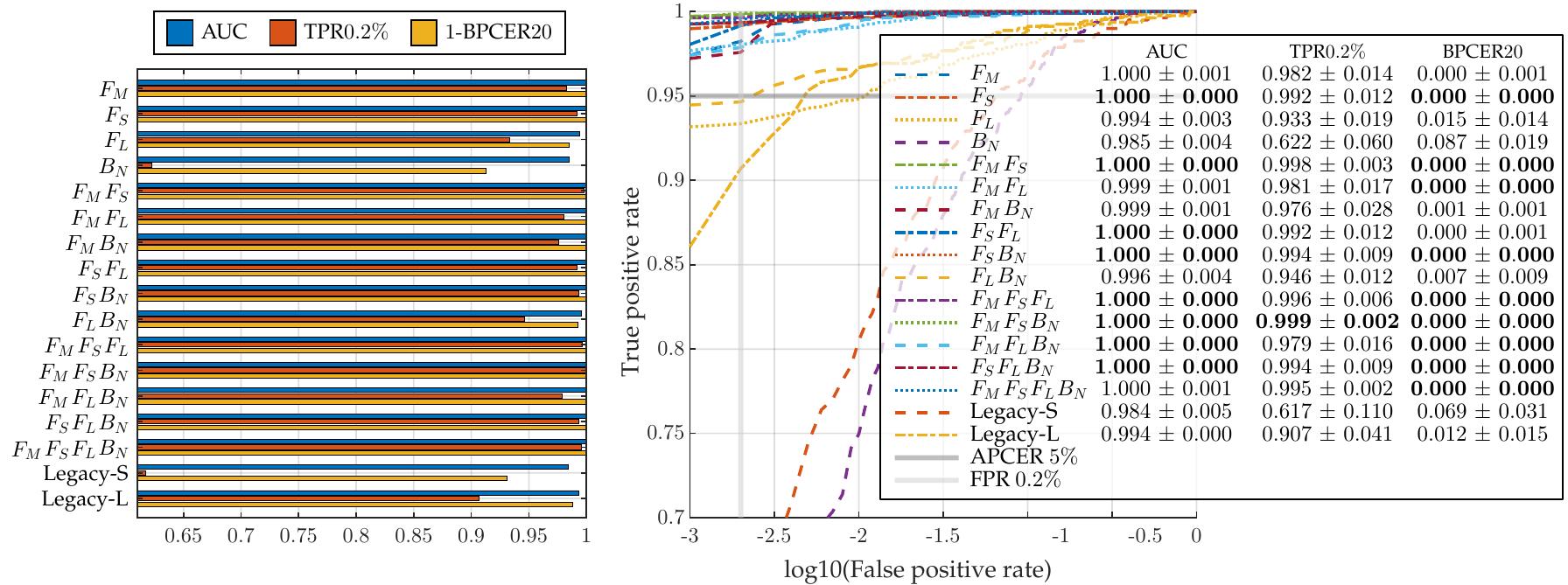}
    \caption{3FOLD FPAD evaluation on the \emph{PADISI-APL} dataset and comparison with performance on legacy data (\emph{APL-LEGACY} from \tabref{tab:padisi_datasets}). Left: Bar-graph visualization of the evaluation metrics (mean for all folds); Right: ROC curves and analytic values (mean and standard deviation for all folds with best values highlighted in bold). Legacy-S refers to ``small" legacy images, equal in size to the used prototype data passed to the network, while Legacy-L refers to ``large" legacy images, extracted as described in the legacy data pre-processing steps of \secref{sec:data_preprocessing}.}
    \label{fig:padisi_apl_3fold}
\end{figure*}

\subsection{Evaluation on \emph{LivDet} dataset}
\label{sec:livdet_results}
We first evaluate the proposed FCN model on the LivDet$2015$ dataset~\cite{LivDet2015}. The dataset contains images from $4$ legacy sensors, namely, CrossMatch, GreenBit, Hi-Scan and Digital-Persona from which, following the pre-processing steps for legacy data, we extracted ROIs of size $320 \times 256$, $320 \times 256$, $600 \times 480$ and $260 \times 200$ pixels on the detected finger area, respectively. Images were also normalized in $[0, 1]$, based on the bit depth of $8$ bits for all sensors. The LivDet$2015$ dataset provides pre-defined training and testing sets for each sensor but no validation set. Therefore, training was performed for the maximum of $100$ epochs using the training parameters described in \secref{sec:method}. The resulting ROC curves and evaluation metrics are depicted in \figref{fig:livdet2015_results} while accuracy comparison with other algorithms using a threshold of $0.5$ is summarized in \tabref{tab:livdet_results}. As observed, the proposed model achieves state-of-the-art performance for all experiments, supporting its power for FPAD.

\subsection{Evaluation on \emph{PADISI} dataset}
\label{sec:padisi_results}
We now evaluate the performance of the FCN model on the \emph{PADISI} dataset. Our analysis uses the pre-processed data cubes and, following an early-fusion approach, considers different stacked combinations of $F_{M}$, $F_{S}$, $F_{L}$ and $B_{N}$ data as input channels to our model under a range of evaluation protocols. Hence for each experiment in each evaluation protocol, we vary the number of channels $C$, as presented in \figref{fig:network}, to consider all possible $15$ combinations of the captured sensing modalities as input data to our model.

\begin{figure*}[!t]
    \centering
    \includegraphics[scale=0.9]{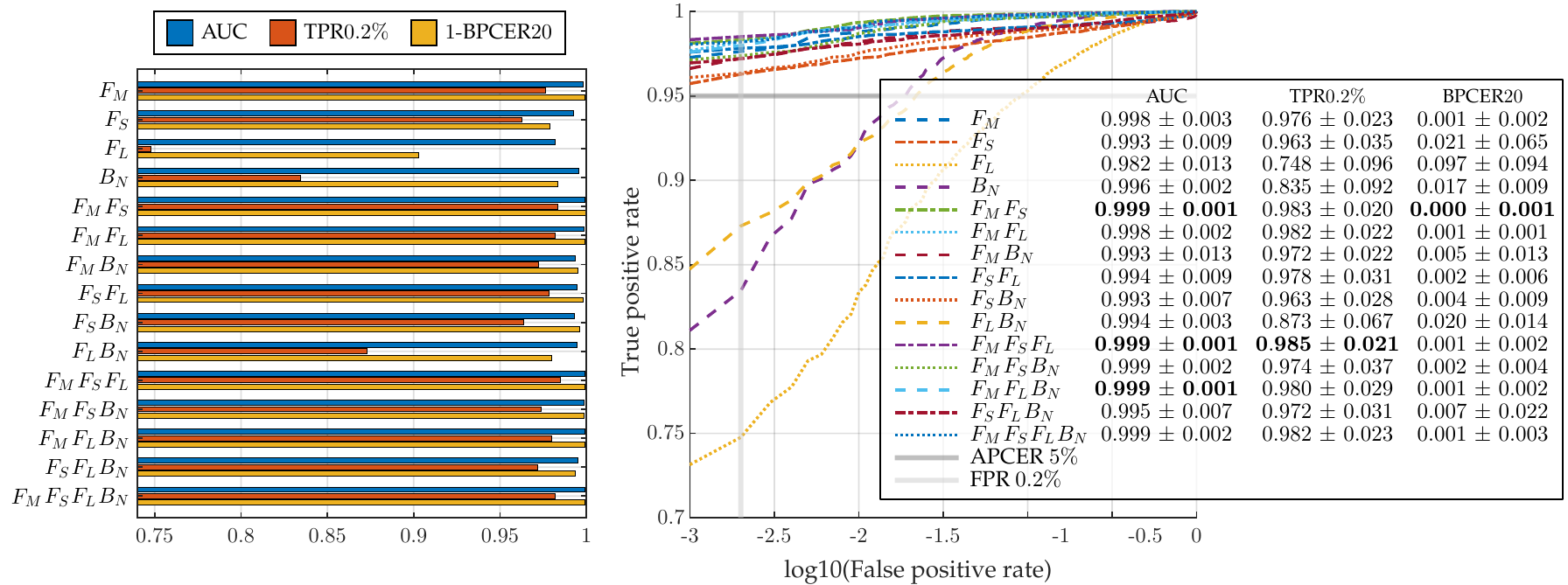}
    \caption{LOO FPAD evaluation on the \emph{PADISI-USC} dataset. Left: Bar-graph visualization of the evaluation metrics (mean for all LOO categories); Right: ROC curves and analytic values (mean and standard deviation for all LOO categories with best values highlighted in bold). See \tabref{tab:padisi_usc_loo_analytic} for analytic results.}
    \label{fig:padisi_usc_loo}
\end{figure*}

\begin{figure*}
    \centering
    \includegraphics[scale=0.9]{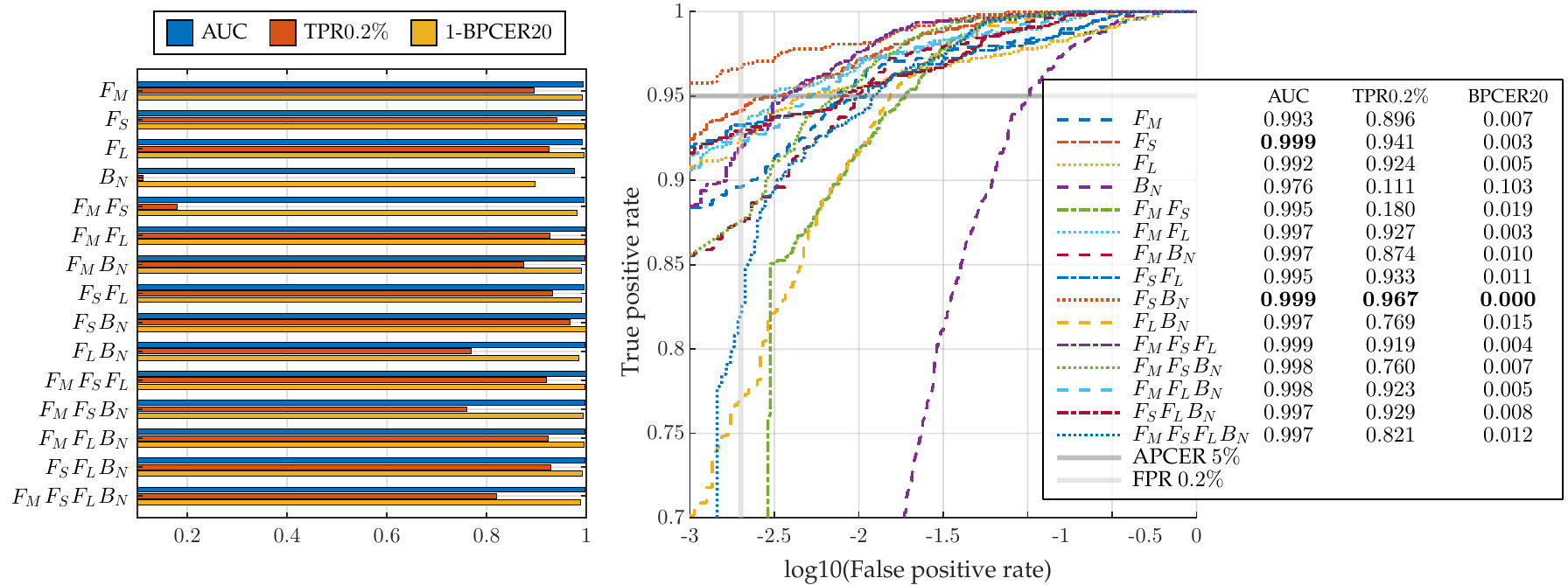}
    \caption{Inter-site FPAD evaluation on the \emph{PADISI} dataset (training on \emph{USC} data and testing on \emph{APL} data). Left: Bar-graph visualization of the evaluation metrics; Right: ROC curves and analytic values (with best values highlighted in bold).}
    \label{fig:cross_dataset}
\end{figure*}

\subsubsection{Evaluation Protocols}
\label{sec:evaluation_protocols}
We employ three different evaluation protocols. In the first strategy, we use a 3-fold (3FOLD) partitioning to alleviate the bias resulting from a fixed division of the dataset into training, testing, and validation sets. In this strategy, samples of a collection are divided into three roughly-equal sets of samples such that data of each participant only appears in a single set. Then, 3FOLD cross-validation evaluation is done by performing three experiments, each time using two sets for training and validation, and the left-out set for testing. Each time, from the training/validation group, $80\%$ of data is used for training and $20\%$ for validation, while making sure all training/testing/validation sets are participant-disjoint. It is also important to note that in all splits, the distribution of bona-fide and different PAI category samples are approximately balanced among the training/testing/validation sets. The 3FOLD partitioning is considered for intra-collection evaluations, which are applied either to the \emph{PADISI-USC} or the \emph{PADISI-APL} collections.

As a second partitioning strategy, we use a leave-one-attack-out (LOO) protocol, in order to evaluate the ability to detect \emph{unknown attacks} (i.e., attacks not present in the training data). Using the PAI grouping attributes of \tabref{tab:PAI_information}, we create a partition for each color (excluding the special case), leading to $11$ partitions for the \emph{PADISI-USC} collection.

As mentioned in Sections \ref{sec:finger_station} and \ref{sec:dataset}, the sensor parameters, environmental settings as well as the demographics for \emph{PADISI} were not identical in the two collection sites (\emph{USC} and \emph{APL}). To study the effect of these variations to the FPAD classification performance, we also perform inter-site evaluation for which \emph{PADISI-USC} is used for training and validation, and \emph{PADISI-APL} is used for testing. 

FPAD classification performance is evaluated, per fold, for each of the settings and for all aforementioned metrics. The mean and standard deviation of each metric is computed over the total number of the folds for each partitioning strategy. In the following sections, we provide an analysis of our results under the different scenarios.

\subsubsection{Evaluation Results}
\label{sec:evaluation_results}

\begin{table*}[!t]
 \caption{Analytic LOO FPAD evaluation, per LOO category, on the \emph{PADISI-USC} dataset. Left: Bar-graph visualization of the TPR$0.2\%$ metric; Right: Analytic values (with best values highlighted in bold). The corresponding average LOO results are presented in \figref{fig:padisi_usc_loo}.}
\label{tab:padisi_usc_loo_analytic}
 \centering
\includegraphics[scale=0.9]{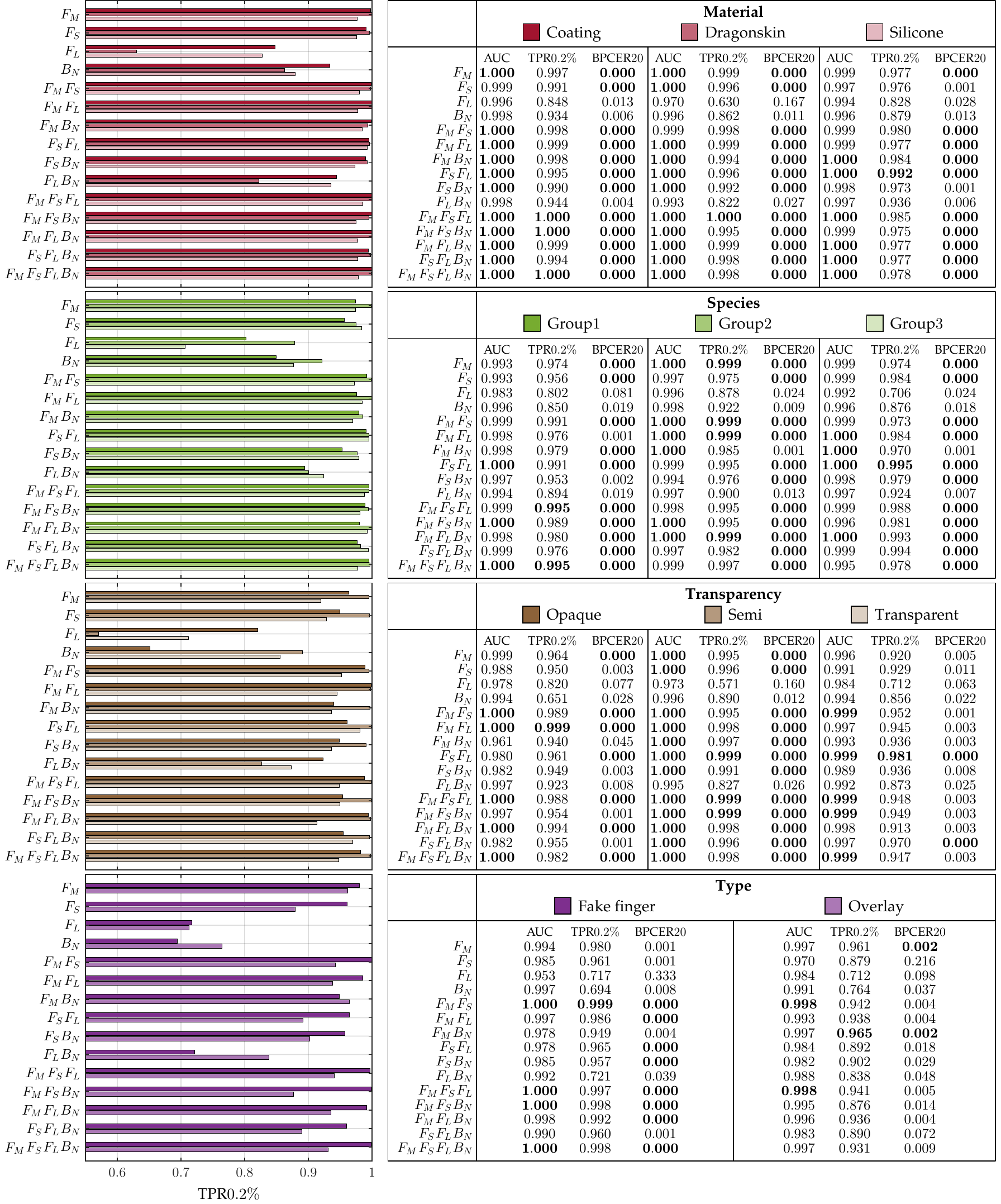}
\end{table*}

\begin{itemize}
    \item \underline{3FOLD evaluation}: The 3FOLD evaluation results for the \emph{PADISI-USC} dataset are presented in \figref{fig:padisi_usc_3fold} and for the \emph{PADISI-APL} in \figref{fig:padisi_apl_3fold}. For the \emph{PADISI-APL} dataset, a comparison to the performance using the data of the selected legacy sensor is presented using two different pre-processing methods. Legacy-S refers to ROIs of $160 \times 80$ pixels, equal in size to our prototype data, while Legacy-L refers to larger ROIs of size $480 \times 384$ following the general pre-processing steps for legacy data described earlier. Even though the total number of samples for the \emph{APL-LEGACY} data is not equal to our prototype data; the folds were consistent by using the same participant's samples as in the 3FOLD partitions described in \secref{sec:evaluation_protocols}.
    \item \underline{\emph{PADISI-USC} LOO evaluation}: The average performance on all $11$ LOO partitions of the \emph{PADISI-USC} dataset is presented in \figref{fig:padisi_usc_loo} while analytic results for each partition are summarized in \tabref{tab:padisi_usc_loo_analytic}.
    \item \underline{Inter-site evaluation}: The inter-site evaluation results are depicted in \figref{fig:cross_dataset}.
\end{itemize}

\subsubsection{Discussion}
\label{sec:padisi_results_discussion}
\begin{figure*}[!t]
    \centering
    \ifarxiv
    \includegraphics{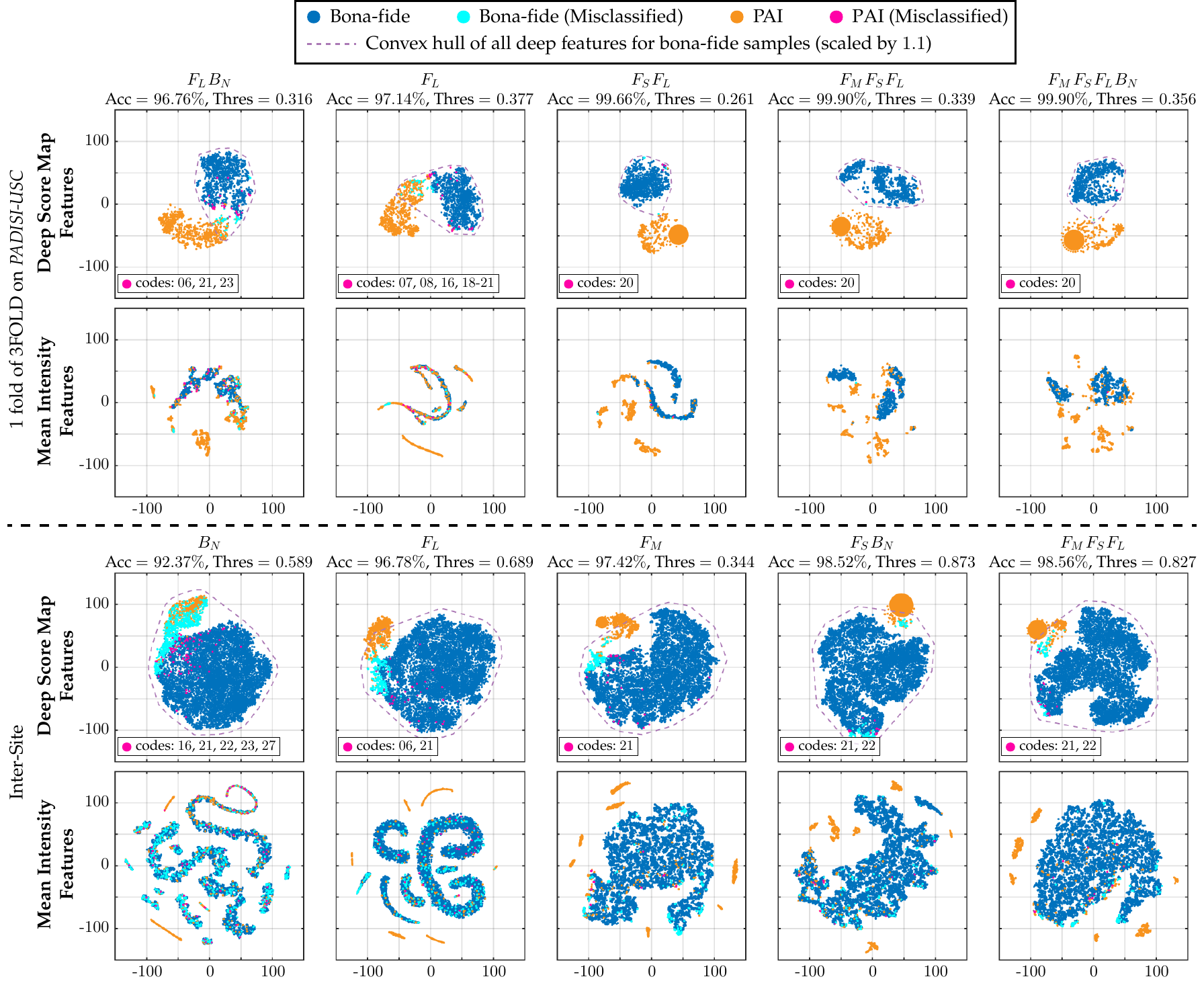}
    \fi
    \caption{t-SNE visualization comparison for deep features (score maps) of the FCN model in \figref{fig:network} and the corresponding mean-intensity features of the utilized input channels. Each experiment presents the classification accuracy based on the EER threshold, the misclassified samples of the FCN model in both visualizations and the codes of misclassified PAIs in the legend. Experiments are sorted by their accuracy from left to right. The reader is referred to \secref{sec:padisi_results_discussion} for a detailed analysis.}
    \label{fig:tsne_mean_intensity_vs_deep_features}
\end{figure*}

Based on the results of \secref{sec:evaluation_results} we can make the following observations:
\begin{itemize}
    \item Some of the studied sensing modalities (particularly $F_{M}$ and $F_{S}$) can achieve very high FPAD performance with $F_{S}$ being less affected by cross-dataset variations (as seen in \figref{fig:padisi_apl_3fold}), which is consistent with the relevant literature on SWIR imaging.
    \item With the exception of $F_{L}$ or $B_{N}$ data alone, any other single sensing modality or combination proves superior to legacy data (see \figref{fig:padisi_apl_3fold}), especially at very low false positive rates.
    \item In the majority of the cases, combining more than one sensing modalities leads to improved performance. This is particularly apparent in the \emph{unknown attack} scenario where most models employing two or more sensing modalities consistently outperform the single ones for most metrics (see \tabref{tab:padisi_usc_loo_analytic}). This observation supports the use of multi-spectral data for FPAD and the power of \emph{hybrid} methods.
    \item In multiple cases, increasing the number of channels provided to the model does not always lead to performance improvement, albeit not with significant loss. Such behavior may signify overfitting or non-optimal weighting of certain channels toward the final classification output and could open new research directions for incorporating channel attention techniques to our model~\cite{Roy2018:Excitation, Woo2018, Zhang2019}.
\end{itemize}
Extending the t-SNE visualization analysis of \secref{sec:dataset}, we compare the separation in t-SNE space between the features extracted by the FCN model (score maps) to the corresponding mean-intensity features, for select experiments, in \figref{fig:tsne_mean_intensity_vs_deep_features}. The illustrations include results for $1$ fold of the 3FOLD evaluation protocol on the \emph{PADISI-USC} collection and the inter-site evaluation protocol. The figure presents the two worst results on the left, the medial result in the middle and the best two results on the right for each evaluation protocol by using the equal error rate (EER) threshold to calculate the accuracy of each experiment. The visualization marks the misclassified bona-fide and PAI samples based on the EER threshold and shows the corresponding locations of these samples in the mean-intensity feature visualization. Finally, the codes of the misclassified PAIs are presented in the legends. This analysis results in the following observations:
\begin{itemize}
    \item The addition of sensing modalities leads to performance improvement, in most cases (compare rightmost to leftmost input channels), consistent with the ROC curve observations.
    \item The mean-intensity feature visualization indeed provides an estimate of how challenging a dataset is. The majority of misclassified PAIs by the FCN model are in most cases within the bona-fide cluster of the mean-intensity t-SNE visualization. This dictates that intensities of multi-spectral data could be playing an important role as classification features in the model.
    \item The majority of misclassified bona-fides lie at the border of the bona-fide cluster in the mean-intensity feature visualization, which further supports the aforementioned argument (this is mostly visible in the inter-site experiment).
    \item The misclassified PAI codes for the rightmost highest accuracy experiments agree with the analysis of \figref{fig:tsne_mean_intensity}, which demonstrated that the most challenging PAIs are the transparent overlays (see \tabref{tab:PAI_information}).
    \item The left-most experiments also demonstrate the ability of deep features to achieve a reasonable classification performance even when large amounts of intensity features are intermingled for the two classes.
    \item Finally, the inter-site analysis clearly shows a shift in the EER threshold resulting from the vastly different demographics and collection system variations for the two datasets. This is consistent with the mean-intensity t-SNE separation of bona-fide samples from different datasets in \figref{fig:tsne_mean_intensity}.
\end{itemize}

\section{Conclusions}
\label{sec:conclusion}
This paper presented a comprehensive analysis of \emph{hybrid} (\emph{hardware-based}) FPAD, using a number of recently introduced sensing modalities, which are front-illuminated visible, NIR, and SWIR images; back-illuminated NIR images, and laser-speckle contrast imaging. The analysis was conducted on a new dataset, named \textit{PADISI-Finger}. \textit{PADISI-Finger} consists of two collections performed at two different sites, \textit{PADISI-USC} and \textit{PADISI-APL}, the former of which will be publicly released upon the acceptance of this manuscript. \textit{PADISI-Finger} contains data from over a thousand participants and covers more than $60$ PAI species. Our analysis employed a novel fully-convolutional network model, whose power was demonstrated by showing its state-of-the-art performance on the LivDet$2015$ dataset. FPAD performance using this FCN model revealed the advantages of some of the individual unconventional sensing modalities and all different combinations of them over legacy data. The power of combining multiple sensing modalities was further confirmed by the results of our rigorous evaluation protocols, which assessed the effects of testing on completely unseen attack categories (leave-one-attack out) as well as under different dataset characteristics (inter-collection). In such challenging protocols, front-illuminated multi-spectral images (visible, NIR, and SWIR) stood out as the most reliable sensing modalities, either individually or in combination with others. Low-dimensional data visualization of the raw average intensity values revealed a notable similarity to the equivalent visualization of the features learned by the FCN model, which upholds the role of the data in the obtained FPAD performance in our analysis. Finally, it was observed that the FPAD performance is not directly correlated with the number of employed sensing modalities. This is believed to be an artifact of the neural network model architecture, and its investigation is deferred to our future work.
\ifCLASSOPTIONcompsoc
  \section*{Acknowledgments}
\else
  \section*{Acknowledgment}
\fi

This research is based upon work supported by the Office of the Director of National Intelligence (ODNI), Intelligence Advanced Research Projects Activity (IARPA), via IARPA R\&D Contract No. 2017-17020200005. The views and conclusions contained herein are those of the authors and should not be interpreted as necessarily representing the official policies or endorsements, either expressed or implied, of the ODNI, IARPA, or the U.S. Government. The U.S. Government is authorized to reproduce and distribute reprints for Governmental purposes notwithstanding any copyright annotation thereon. 

The authors would also like to acknowledge the efforts of the APL team for organizing and managing the large scale data collection at their site in Columbia, Maryland, and the effort of our collaborators from Hochschule Darmstadt – University of Applied Sciences for their help and support in collecting fingerprint data at the APL site. A final acknowledgement is due to Mian Zhong and Hanchen Xie for their efforts towards a very early draft of this manuscript.

\ifCLASSOPTIONcaptionsoff
  \newpage
\fi
\bibliographystyle{IEEEtran}
\bibliography{biblio}
%

\begin{IEEEbiography}[{\includegraphics[]{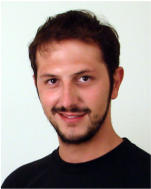}}]{Leonidas Spinoulas}
Leonidas Spinoulas received his Diploma degree in Electrical and Computer Engineering from the National Technical University of Athens, Greece in 2010. In September 2010 he joined Northwestern University, Evanston, IL, USA and the Image and Video Processing Laboratory (IVPL) under the supervision of Prof. Aggelos K. Katsaggelos. He received the M. Sc. Degree in Electrical Engineering and Computer Science in 2012 and the Ph.D. degree from the same department in August 2016. Since 2017, he holds a Research Computer Scientist position with the Information Sciences Institute (University of Southern California), Marina del Rey, CA. He was previously a Research Scientist for Ricoh Innovations Corporation, Cupertino, CA, USA. He was the recipient of the best paper awards at EUSIPCO 2013 and SENSORCOMM 2015 and has 4 patents. His primary research interests include deep learning, biometrics, multispectral imaging, image processing, image restoration, inverse problems and compressive sensing.
\end{IEEEbiography}

\begin{IEEEbiography}[{\includegraphics[]{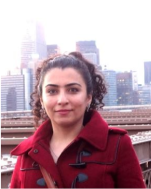}}]{Hengameh Mirzaalian}
Hengameh Mirzaalian conducts research on image-processing, machine-learning, and deep-learning techniques to address different tasks on images with different modalities and applications. She received her Ph.D. in computing science from Simon Fraser University, BC. Prior to ISI, she worked with various research institutes including Siemens Corporate Research and Technologies, Harvard Medical School, Boston Children’s Hospital, Brigham and Women’s Hospital and Nasa Jet Propulsion Laboratory. Her research appeared on top journals including Medical Image Analysis, NeuroImage and IEEE conferences.
\end{IEEEbiography}

\begin{IEEEbiography}[{\includegraphics[]{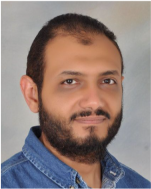}}]{Mohamed Hussein}
Dr. Mohamed E. Hussein is a computer scientist at USC ISI and an associate professor (on leave) at Alexandria University, Egypt. Dr. Hussein obtained his Ph.D. degree in Computer Science from the University of Maryland at College Park, MD, USA in 2009. Then, he spent close to two years as an Adjunct Member Research Staff at Mitsubishi Electric Research Labs, Cambridge, MA, before moving to Alexandria University as a faculty member. Prior to joining ISI, he spent three years at Egypt-Japan University of Science and Technology (E-JUST), in Alexandria, Egypt. Dr. Hussein's most recent research focus has been in securing biometrics and machine learning systems. He has over 30 published papers, and three issued patents.
\end{IEEEbiography}

\begin{IEEEbiography}[{\includegraphics[]{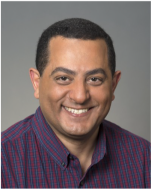}}]{Wael AbdAlmageed}
Dr. AbdAlmageed is a Research Associate Professor at the Department of Electrical and Computer Engineering, and a research Team Leader and Supervising Computer Scientist with Information Sciences Institute, both being units of USC Viterbi School of Engineering. His research interests include representation learning, debiasing and fair representations, multimedia forensics and visual misinformation (including deepfake and image manipulation detection) and biometrics. Prior to joining ISI, Dr. AbdAlmageed was a research scientist with the University of Maryland at College Park, where he led several research efforts for various NSF, DARPA and IARPA programs. He obtained his Ph.D. with Distinction from the University of New Mexico in 2003 where he was also awarded the Outstanding Graduate Student award. He has two patents and over 70 publications in top computer vision and high performance computing conferences and journals. Dr. AbdAlmageed is the recipient of 2019 USC Information Sciences Institute Achievement Award.
\end{IEEEbiography}





\end{document}